%% file: blei11.tex
\documentclass[12pt,preprint]{imsart}

\input{preamble.tex}

\begin{document}

\begin{frontmatter}

\title{Distance Dependent Chinese Restaurant Processes}
\runtitle{Distance Dependent Chinese Restaurant Processes}

\begin{aug}
\author{\fnms{David M.} \snm{Blei}\ead[label=e1]{blei@cs.princeton.edu}}

\runauthor{Blei and Frazier}

\address{Department of Computer Science \\ Princeton University\\
  Princeton, NJ, USA \\ \printead{e1}}

\author{\fnms{Peter I.} \snm{Frazier}\ead[label=e2]{pf98@cornell.edu}}

\address{School of Operations Research and Information
  Engineering \\ Cornell University \\ Ithaca, NY 14853, USA \\ \printead{e2}}
\end{aug}
\end{frontmatter}





\maketitle

\begin{abstract}
  We develop the distance dependent Chinese restaurant process, a
  flexible class of distributions over partitions that allows for
  dependencies between the elements.  This class can be used to model
  many kinds of dependencies between data in infinite clustering
  models, including dependencies arising from time, space, and network
  connectivity.  We examine the properties of the distance dependent
  CRP, discuss its connections to Bayesian nonparametric mixture
  models, and derive a Gibbs sampler for both fully observed and
  latent mixture settings.  We study its empirical performance with
  three text corpora.  We show that relaxing the assumption of
  exchangeability with distance dependent CRPs can provide a better
  fit to sequential data and network data.  We also show that the
  distance dependent CRP representation of the traditional CRP mixture
  leads to a faster-mixing Gibbs sampling algorithm than the one based
  on the original formulation.
\end{abstract}


\input{intro.tex}
\input{ddcrp.tex}
\input{inference-alt.tex}

\input{marg-invariance.tex}
\input{study.tex}

\input{discussion.tex}

\bibliographystyle{apalike}
\bibliography{bib}

\newpage
\appendix

\input{proofs-alt.tex}

\end{document}

%% file: preamble.tex
\newcommand{\mysec}[1]{Section~\ref{sec:#1}}

\newcommand{\myeq}[1]{Eq.~(\ref{eq:#1})}

\newcommand{\g}{\,\vert\,}

\newcommand{\R}{\mathbb{R}}

\newcommand{\bc}{\mathbf{c}}
\newcommand{\bx}{\mathbf{x}}

\usepackage{times} 
\usepackage{amsmath}
\usepackage{amssymb}
\usepackage{amsfonts}
\usepackage{epsfig}
\usepackage{url}
\usepackage{rotating}

\usepackage{microtype}

\usepackage{graphicx}

\newcommand{\myfig}[1]{Figure~\ref{fig:#1}}
\newcommand{\ind}[1]{1[#1]}
\newcommand{\rmnew}{\textrm{new}}

\newcommand{\Prob}[1]{\mathbb{P}\left\{#1\right\}}
\newcommand{\X}{\mathbb{X}}
\newcommand{\Y}{\mathbb{Y}}
\newcommand{\Pmeasure}{\mathbb{P}}

\newenvironment{packed_enumerate}{
  \begin{enumerate}
    \setlength{\topsep}{0pt}
    \setlength{\itemsep}{6pt}
    \setlength{\parskip}{0pt}
    \setlength{\parsep}{0pt}
}{\end{enumerate}}

\usepackage{natbib}
\usepackage{amsthm}
\newtheorem{proposition}{Proposition}
\newtheorem{corollary}{Corollary}
\usepackage[left=1in, right=1in, top=1in, bottom=1in]{geometry}


%% file: intro.tex
\section{Introduction}

Dirichlet process (DP) mixture models provide a valuable suite of
flexible clustering algorithms for high dimensional data analysis.
Such models have been adapted to text
modeling~\citep{Teh:2007,Goldwater:2006a}, computer
vision~\citep{Sudderth:2005}, sequential
models~\citep{Dunson:2006,Fox:2007}, and computational
biology~\citep{Xing:2007}.  Moreover, recent years have seen
significant advances in scalable approximate posterior inference
methods for this class of
models~\citep{Liang:2007a,Daume:2007,Blei:2005}.  DP mixtures have
become a valuable tool in modern machine learning.

DP mixtures can be described via the Chinese restaurant process (CRP),
a distribution over partitions that embodies the assumed prior
distribution over cluster structures~\citep{Pitman:2002}.  The CRP is
fancifully described by a sequence of customers sitting down at the
tables of a Chinese restaurant.  Each customer sits at a previously
occupied table with probability proportional to the number of
customers already sitting there, and at a new table with probability
proportional to a concentration parameter.  In a CRP mixture,
customers are identified with data points, and data sitting
at the same table belong to the same cluster.  Since the number of
occupied tables is random, this provides a flexible model in which the
number of clusters is determined by the data.

The customers of a CRP are exchangeable---under any permutation of
their ordering, the probability of a particular configuration is the
same---and this property is essential to connect the CRP mixture to
the DP mixture.  The reason is as follows.  The Dirichlet process is a
distribution over distributions, and the DP mixture assumes that the
random parameters governing the observations are drawn from a
distribution drawn from a Dirichlet process.  The observations are
conditionally independent given the random distribution, and thus they
must be marginally exchangeable.\footnote{That these parameters will
  exhibit a clustering structure is due to the discreteness of
  distributions drawn from a Dirichlet
  process~\citep{Ferguson:1973,Antoniak:1974,Blackwell:1973a}.}  If
the CRP mixture did not yield an exchangeable distribution, it could
not be equivalent to a DP mixture.

Exchangeability is a reasonable assumption in some clustering
applications, but in many it is not.  Consider data ordered in time,
such as a time-stamped collection of news articles.  In this setting,
each article should tend to cluster with other articles that are
nearby in time.  Or, consider spatial data, such as pixels in an image
or measurements at geographic locations.  Here again, each datum should
tend to cluster with other data that are nearby in space.  While the
traditional CRP mixture provides a flexible prior over partitions of
the data, it cannot accommodate such non-exchangeability.

In this paper, we develop the \textit{distance dependent Chinese
  restaurant process}, a new CRP in which the random seating
assignment of the customers depends on the distances between
them.\footnote{This is an expanded version of our shorter conference
  paper on this subject~\citep{Blei:2010a}.  This version contains new
  perspectives on inference and new results.} These distances can be
based on time, space, or other characteristics.  Distance dependent
CRPs can recover a number of existing dependent
distributions~\citep{Ahmed:2008,Zhu:2005}.  They can also be arranged
to recover the traditional CRP distribution.  The distance dependent
CRP expands the palette of infinite clustering models, allowing for
many useful non-exchangeable distributions as priors on
partitions.\footnote{We avoid calling these clustering models
  ``Bayesian nonparametric'' (BNP) because they cannot necessarily be
  cast as a mixture model originating from a random measure, such as
  the DP mixture model.  The DP mixture is BNP because it includes a
  prior over the infinite space of probability densities, and the CRP
  mixture is only BNP in its connection to the DP mixture.  That said,
  most applications of this machinery are based around letting the
  data determine their number of clusters.  The fact that it actually
  places a distribution on the infinite-dimensional space of
  probability measures is usually not exploited.}

The key to the distance dependent CRP is that it represents the
partition with \textit{customer assignments}, rather than table
assignments.  While the traditional CRP connects customers to tables,
the distance dependent CRP connects customers to other customers.  The
partition of the data, i.e., the table assignment representation,
arises from these customer connections.  When used in a Bayesian
model, the customer assignment representation allows for a
straightforward Gibbs sampling algorithm for approximate posterior
inference (see \mysec{inference}).  This provides a new tool for
flexible clustering of non-exchangeable data, such as time-series or
spatial data, as well as a new algorithm for inference with
traditional CRP mixtures.


\paragraph{Related work.} Several other non-exchangeable priors on
partitions have appeared in recent research literature.  Some can be
formulated as distance dependent CRPs, while others represent a
different class of models.  The most similar to the distance dependent
CRP is the probability distribution on partitions presented in
\cite{Dahl:2008}.  Like the distance dependent CRP, this distribution
may be constructed through a collection of independent priors on
customer assignments to other customers, which then implies a prior on
partitions.  Unlike the distance dependent CRP, however, the
distribution presented in \cite{Dahl:2008} requires 
normalization of these customer assignment probabilities.  The
model in \cite{Dahl:2008} may always be written as a distance
dependent CRP, although the normalization requirement prevents
the reverse from being true (see \mysec{model}).  We note that
\cite{Dahl:2008} does not present an algorithm for sampling from the
posterior, but the Gibbs sampler presented here for the distance
dependent CRP can also be employed for posterior inference in that model.


There are a number of Bayesian nonparametric models that allow for
dependence between (marginal) partition membership probabilities.
These include the dependent Dirichlet process \citep{MacEachern:1999}
and other similar processes~\citep{Duan:2006,Griffin:2006,Xue:2007}.
Such models place a prior on collections of sampling distributions
drawn from Dirichlet processes, with one sampling distribution drawn
per possible value of covariate and sampling distributions from
similar covariates more likely to be similar.  Marginalizing out the
sampling distributions, these models induce a prior on partitions by
considering two customers to be clustered together if their sampled
values are equal.  (Recall, these sampled values are drawn from the
sampling distributions corresponding to their respective covariates.)
This prior need not be exchangeable if we do not condition on the
covariate values.

Distance dependent CRPs represent an alternative strategy for modeling
non-exchangeability.  The difference hinges on \text{marginal
  invariance}, the property that a missing observation does not affect
the joint distribution.  In general, dependent DPs exhibit marginal
invariance while distance dependent CRPs do not.  For the
practitioner, this property is a modeling choice, which we discuss in
\mysec{model}.  \mysec{marg} shows that distance dependent
CRPs and dependent DPs represent nearly distinct classes of models,
intersecting only in the original DP or CRP.

Still other prior distributions on partitions include those presented
in \cite{Ahmed:2008} and \cite{Zhu:2005}, both of which are special
cases of the distance dependent CRP.  
\cite{Rasmussen:2002} use a gating network similar to the distance dependent CRP
to partition datapoints among experts in way that is more likely to assign nearby points to the same cluster.
Also included are the product
partition models of \cite{Hartigan:1990}, their recent extension to
dependence on covariates \citep{MuellerPPM:2008}, and the dependent
Pitman-Yor process \citep{Sudderth:2008}.  A review of prior probability
distributions on partitions is presented in \cite{MuellerReview:2008}.
The Indian Buffet Process, a Bayesian non-parametric prior on sparse binary
matrices, has also been generalized to model non-exchangeable data by
\cite{Miller:2008}.  We further discuss these priors in relation to the
distance dependent CRP in \mysec{model}.


\vspace{0.15in} The rest of this paper is organized as follows.  In
\mysec{model} we develop the distance dependent CRP and discuss its
properties.  We show how the distance dependent CRP may be used to
model discrete data, both fully-observed and as part of a mixture
model.  In \mysec{inference} we show how the customer assignment
representation allows for an efficient Gibbs sampling algorithm.  In
\mysec{marg} we show that distance dependent CRPs and dependent DPs
represent distinct classes of models.  Finally, in \mysec{study} we
describe an empirical study of three text corpora using the distance
dependent CRP.  We show that relaxing the assumption of
exchangeability with distance dependent CRPs can provide a better fit
to sequential data.  We also show its alternative formulation of the
traditional CRP leads to a faster-mixing Gibbs sampling algorithm than
the one based on the original formulation.


%% file: ddcrp.tex
\section{Distance dependent CRPs}
\label{sec:model}

\begin{figure}[t]
  \begin{center}
    \includegraphics[width=\textwidth]{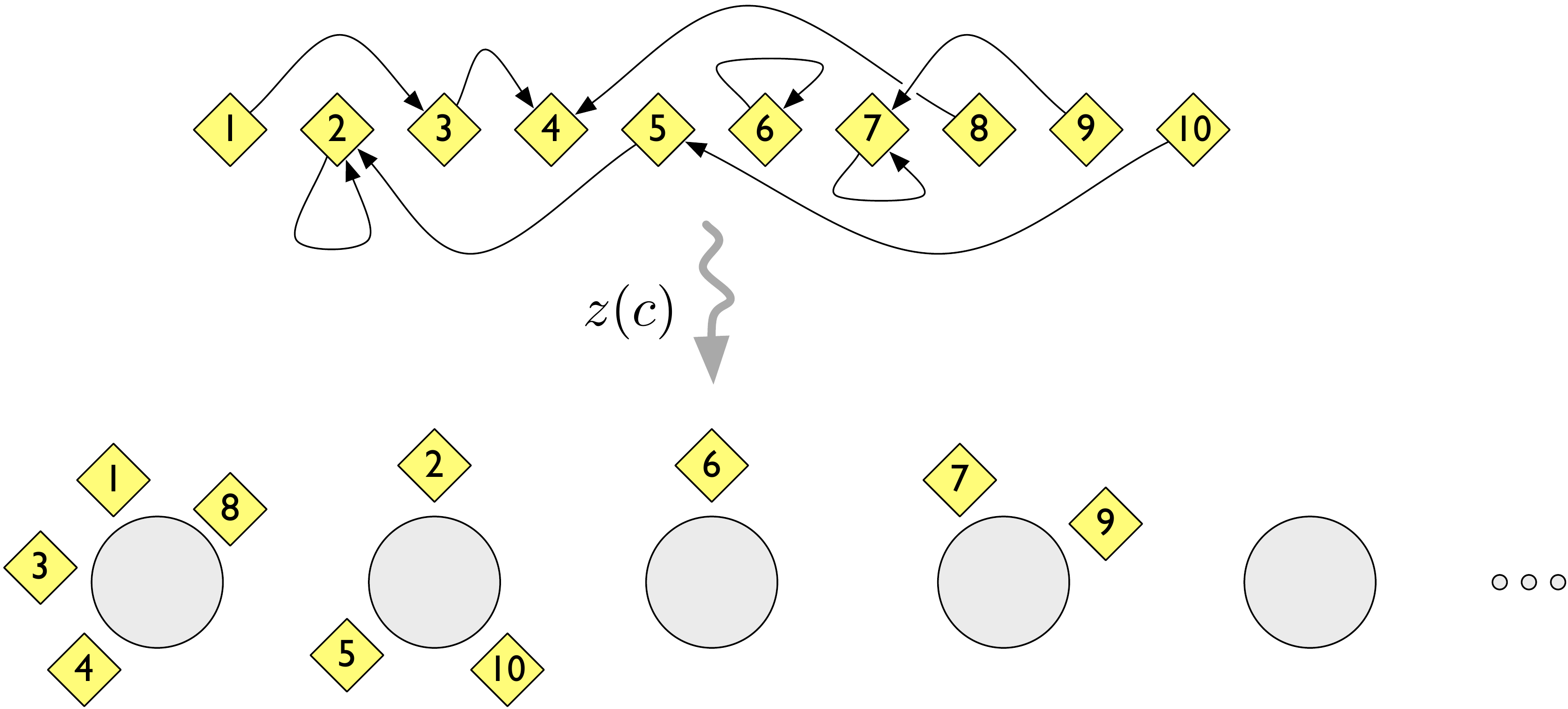}
  \end{center}
  \caption{An illustration of the distance dependent CRP.  The process
    operates at the level of customer assignments, where each customer
    chooses either another customer or no customer according to
    \myeq{dec-crp-prior}.  Customers that chose not to connect to
    another are indicated with a self link The table assignments, a
    representation of the partition that is familiar to the CRP, are
    derived from the customer assignments.\label{fig:tables}}
\end{figure}

\begin{figure*}[htp]
  \begin{center}
  \begin{tabular}{cc}
    \includegraphics[width=0.5\textwidth]{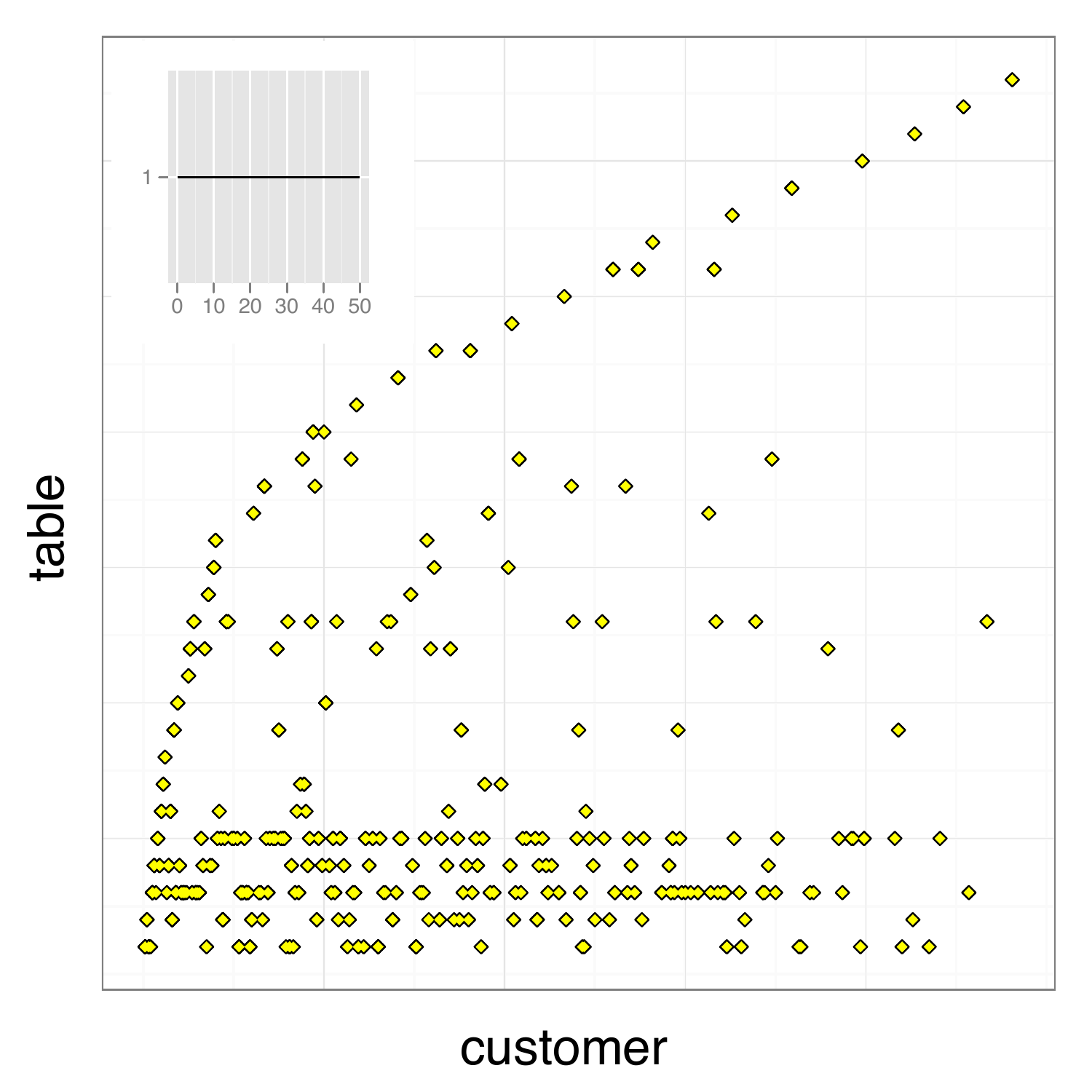} &
    \includegraphics[width=0.5\textwidth]{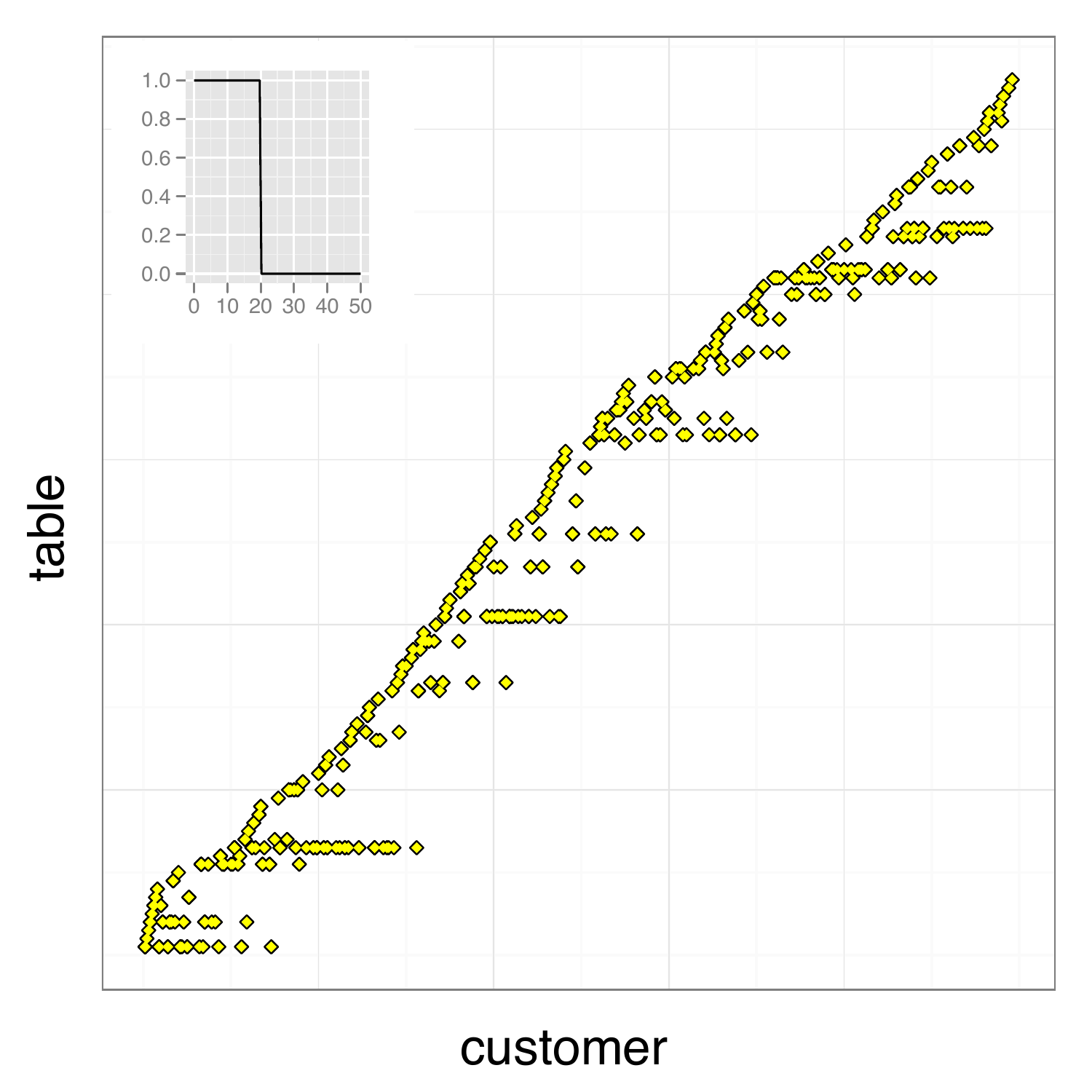} \\
    \includegraphics[width=0.5\textwidth]{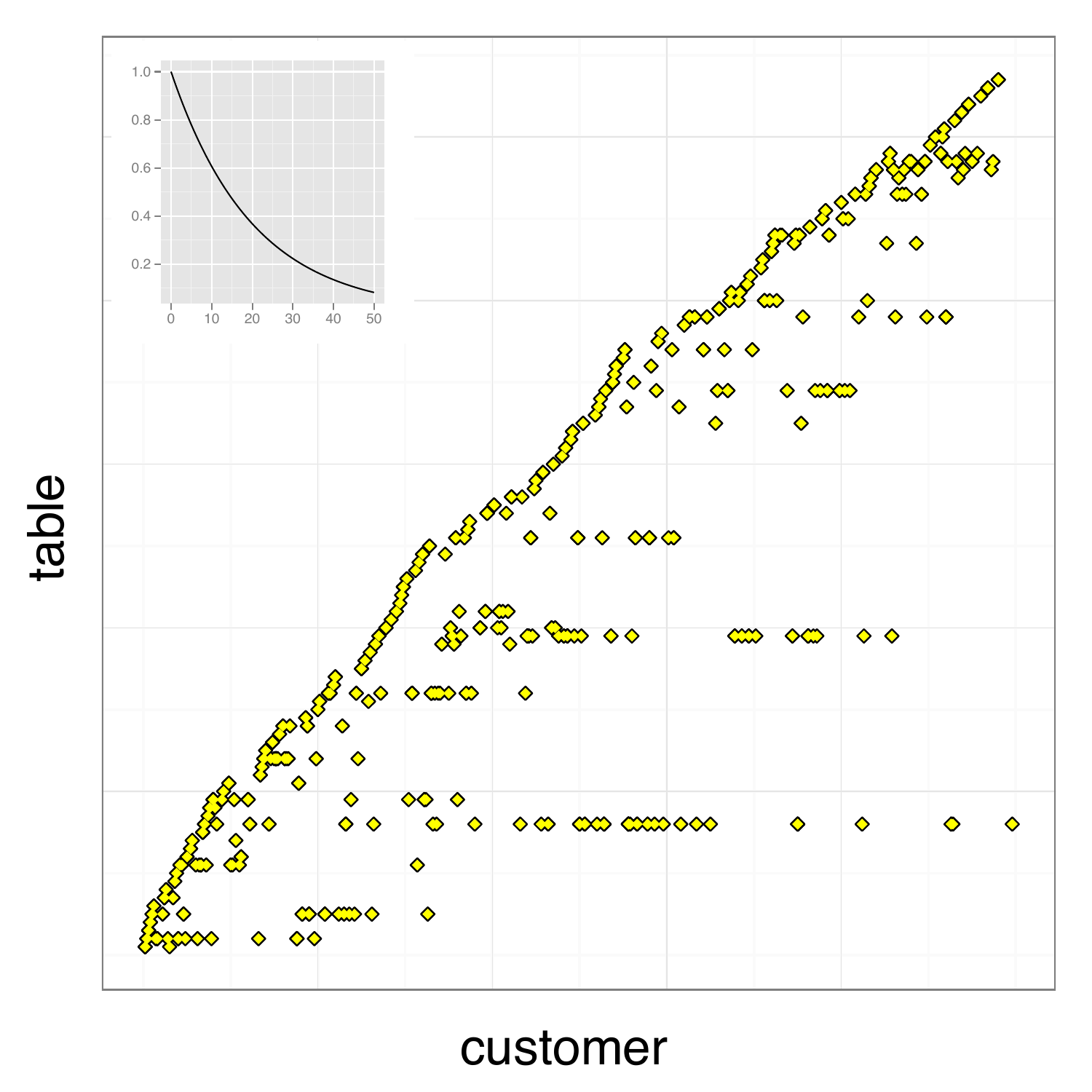} &
    \includegraphics[width=0.5\textwidth]{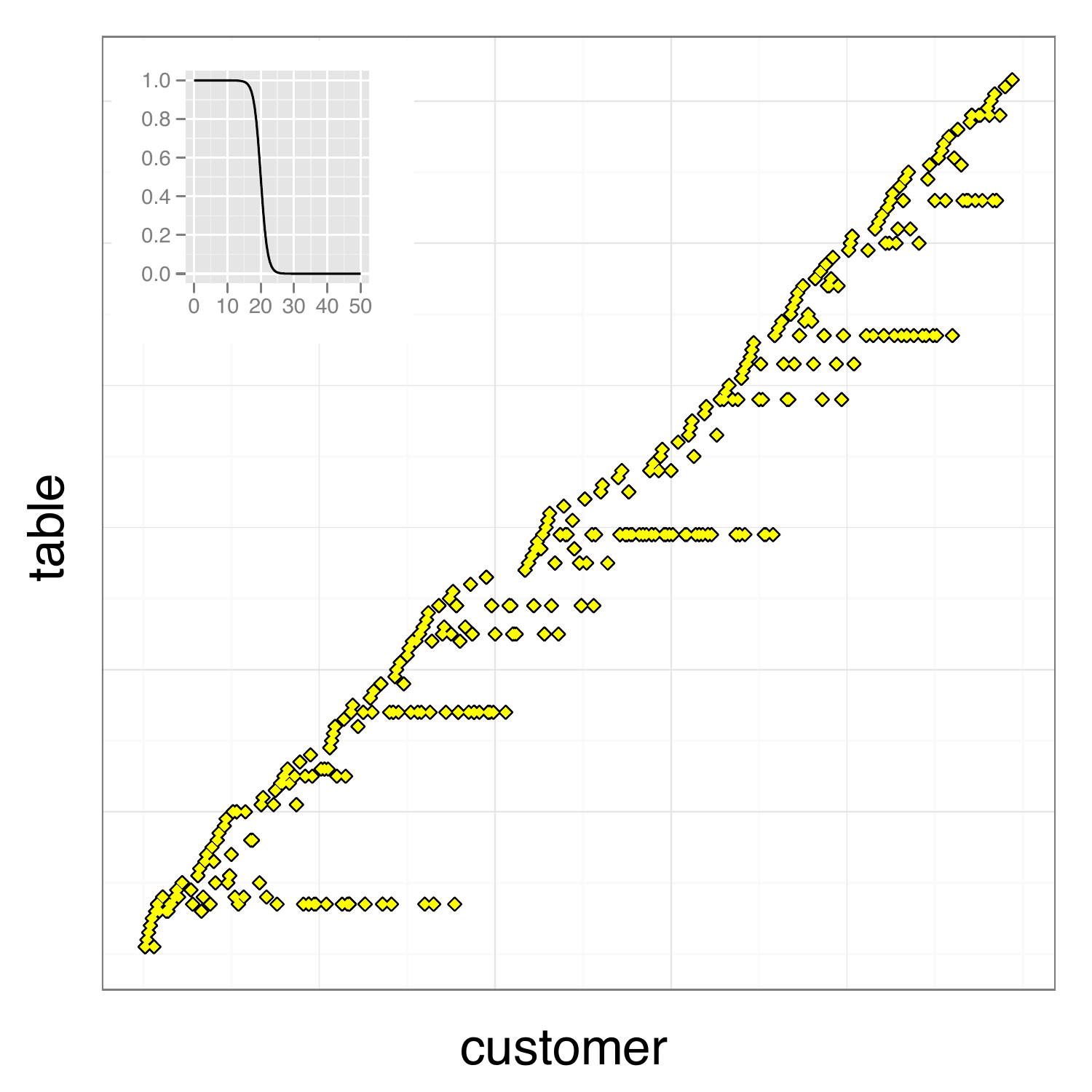}
  \end{tabular}
\end{center}
\caption{Draws from sequential CRPs.  Illustrated are draws for
  different decay functions, which are inset: (1) The traditional CRP;
  (2) The window decay function; (3) The exponential decay function;
  (4) The logistic decay function.  The table assignments
  are illustrated, which are derived from the customer assignments drawn from
  the distance dependent CRP.  The decay functions (inset) are
  functions of the distance between the current customer and each
  previous customer. \label{fig:priors}}
\end{figure*}

The Chinese restaurant process (CRP) is a probability distribution over
partitions~\citep{Pitman:2002}.  It is described by considering a
Chinese restaurant with an infinite number of tables and a sequential
process by which customers enter the restaurant and each sit down at a
randomly chosen table.  After $N$ customers have sat down, their
configuration at the tables represents a random partition.  Customers
sitting at the same table are in the same cycle.

In the traditional CRP, the probability of a customer sitting at a
table is computed from the number of other customers already sitting
at that table.  Let $z_i$ denote the table assignment of the $i$th
customer, assume that the customers $z_{1:(i-1)}$ occupy $K$ tables,
and let $n_k$ denote the number of customers sitting at table $k$.
The traditional CRP draws each $z_i$ sequentially,
\begin{equation}
  p(z_i = k \g z_{1:(i-1)}, \alpha) \propto
  \left\{
    \begin{array}{ll}
      n_k & \textrm{for} \quad k \leq K\\
      \alpha & \textrm{for} \quad k=K+1,
    \end{array}
  \right.
  \label{eq:crp-prior}
\end{equation}
where $\alpha$ is a given scaling parameter.
When all $N$ customers have been seated, their table assignments
provide a random partition.  Though the process is described
sequentially, the CRP is exchangeable.  The probability of a
particular partition of $N$ customers is invariant to the order in
which they sat down.

We now introduce the \textit{distance dependent CRP}.  In this
distribution, the seating plan probability is described in terms of
the probability of a customer sitting with each of the other
\textit{customers}.  The allocation of customers to tables is a
by-product of this representation.  If two customers are reachable by
a sequence of interim customer assignments, then they at the same
table.  This is illustrated in \myfig{tables}.

Let $c_i$ denote the $i$th customer assignment, the index of the
customer with whom the $i$th customer is sitting.  Let $d_{ij}$ denote
the distance measurement between customers $i$ and $j$, let $D$ denote
the set of all distance measurements between customers, and let $f$ be
a decay function (described in more detail below).  The distance
dependent CRP independently draws the customer assignments conditioned
on the distance measurements,
\begin{equation}
  p(c_i = j \g D, \alpha) \propto
  \left\{
    \begin{array}{ll}
      f(d_{ij}) & \textrm{if} \quad j \neq i \\
      \alpha & \textrm{if} \quad i = j.
    \end{array}
  \right.
  \label{eq:dec-crp-prior}
\end{equation}
Notice the customer assignments do not depend on other customer
assignments, only the distances between customers.  Also notice that
$j$ ranges over the entire set of customers, and so any customer may
sit with any other.  (If desirable, restrictions are possible through
the distances $d_{ij}$. See the discussion below of sequential CRPs.)

As we mentioned above, customers are assigned to tables by considering
sets of customers that are reachable from each other through the
customer assignments.  (Again, see \myfig{tables}.)  We denote the
induced table assignments $z(\bc)$, and notice that many
configurations of customer assignments $\bc$ might lead to the same
table assignment.  Finally, customer assignments can produce a cycle,
e.g., customer 1 sits with 2 and customer 2 sits with 1.  This still
determines a valid table assignment: All customers sitting in a cycle
are assigned to the same table.

By being defined over customer assignments, the distance dependent CRP
provides a more expressive distribution over partitions than models
based on table assignments.  This distribution is determined by the
nature of the distance measurements and the decay function.  For
example, if each customer is time-stamped, then $d_{ij}$ might be the
time difference between customers $i$ and $j$; the decay function can
encourage customers to sit with those that are contemporaneous.  If
each customer is associated with a location in space, then $d_{ij}$
might be the Euclidean distance between them; the decay function can
encourage customers to sit with those that are in
proximity.\footnote{The probability distribution over partitions
  defined by \myeq{dec-crp-prior} is similar to the distribution over
  partitions presented in \cite{Dahl:2008}.  That probability
  distribution may be specified by \myeq{dec-crp-prior} if $f(d_{ij})$
  is replaced by a non-negative value $h_{ij}$ that satisfies a
  normalization requirement $\sum_{i\ne j} h_{ij} = N-1$ for each $j$.
  Thus, the model presented in \cite{Dahl:2008} may be understood as a
  normalized version of the distance dependent CRP.  To write this
  model as a distance dependent CRP, take $d_{ij} = 1/h_{ij}$ and
  $f(d) = 1/d$ (with $1/0=\infty$ and $1/\infty=0$), so that
  $f(d_{ij})=h_{ij}$.}  For many sets of distance measurements, the
resulting distribution over partitions is no longer exchangeable; this
is an appropriate distribution to use when exchangeability is not a
reasonable assumption.


\paragraph{Decay functions.}  In general, the decay function
mediates how distances between customers affect the resulting
distribution over partitions.  We assume that the decay function $f$ is
non-increasing, takes non-negative finite values, and satisfies
$f(\infty)=0$.  We consider several types of decay as examples, all of which
satisfy these nonrestrictive assumptions.

The \textit{window decay} $f(d) = 1[d<a]$ only considers
customers that are at most distance $a$ from the current customer. The
\textit{exponential decay} $f(d) = e^{-d/a}$ decays the probability of
linking to
an earlier customer exponentially with the distance to the current
customer. The \textit{logistic decay} $f(d) = \exp{(-d + a)} / (1 +
\exp{(-d + a)})$ is a smooth version of the window decay.  Each of these
affects the distribution over partitions in a different way.

\paragraph{Sequential CRPs and the traditional CRP.}  With certain
types of distance measurements and decay functions, we obtain the
special case of \textit{sequential CRPs}.\footnote{Even though the
  traditional CRP is described as a sequential process, it gives an
  exchangeable distribution.  Thus, sequential CRPs, which include
  both the traditional CRP as well as non-exchangeable distributions,
  are more expressive than the traditional CRP.}  A sequential CRP is
constructed by assuming that $d_{ij} = \infty$ for those $j>i$.  With
our previous requirement that $f(\infty) = 0$, this guarantees that no
customer can be assigned to a later customer, i.e., $p(c_i \leq i \g
D) = 1$.  The sequential CRP lets us define alternative formulations
of some previous time-series models.  For example, with a window decay
function and $a=1$, we recover the model studied in~\cite{Ahmed:2008}.
With a logistic decay function, we recover
the model studied in~\cite{Zhu:2005}.  In our empirical study we will
examine sequential models in detail.

The sequential CRP can re-express the traditional CRP.  Specifically,
the traditional CRP is recovered when $f(d) = 1$ for $d \neq \infty$
and $d_{ij}<\infty$ for $j < i$. To see this, consider the marginal
distribution of a customer sitting at a particular table, given the
previous customers' assignments.  The probability of being assigned to
each of the other customers at that table is proportional to one.
Thus, the probability of sitting at that table is proportional to the
number of customers already sitting there.  Moreover, the probability
of not being assigned to a previous customer is proportional to the
scaling parameter $\alpha$.  This is precisely the traditional CRP
distribution of \myeq{crp-prior}.  Although these models are
the same, the corresponding Gibbs samplers are different (see
\mysec{crp-comparison}).

\myfig{priors} illustrates seating assignments (at the \textit{table}
level) derived from draws from sequential CRPs with each of the decay
functions described above, including the original CRP.  (To adapt
these settings to the sequential case, the distances are $d_{ij} =
i - j$ for $j < i$ and $d_{ij}=\infty$ for $j > i$.)
Compared to the traditional CRP, customers tend to sit at the same
table with other nearby customers.  We emphasize that sequential CRPs
are only one type of distance dependent CRP.  Other distances,
combined with the formulation of \myeq{dec-crp-prior}, lead to a
variety of other non-exchangeable distributions over partitions.

\paragraph{Marginal invariance.}  The traditional CRP is
\textit{marginally invariant}: Marginalizing over a particular
customer gives the same probability distribution as if that customer
were not included in the model at all.  The distance dependent CRP
does not generally have this property, allowing it to capture the way
in which influence might be transmitted from one point to another.
See \mysec{marg} for a precise characterization of the class of
distance dependent CRPs that are marginally invariant.



To see when this might be a relevant property, consider the goal of
modeling preferences of people within a social network.  The model
used should reflect the fact that persons A and B are more likely to
share preferences if they also share a common friend C.  Any
marginally invariant model, however, would insist that the
distribution of the preferences of A and B is the same whether (1)
they have no such common friend C, or (2) they do but his preferences
are unobserved and hence marginalized out.  In this setting, we might
prefer a model that is not marginally invariant.  Knowing that they
have a common friend affects the probability that A and B share
preferences, regardless of whether the friend's preferences are
observed.  A similar example is
modeling the spread of disease.  Suddenly discovering a city between
two others--even if the status of that city is unobserved--should
change our assessment of the probability that the disease travels
between them.

We note, however, that if observations are missing then models that
are not marginally invariant require that relevant conditional
distributions be computed as ratios of normalizing constants.  In
contrast, marginally invariant models afford a more convenient
factorization, and so allow easier computation.  Even when faced with
data that clearly deviates from marginal invariance, the modeler may
be tempted to use a marginally invariant model, choosing computational
convenience over fidelity to the data.

\vspace{0.15in} We have described a general formulation of the
distance dependent CRP.  We now describe two applications to Bayesian
modeling of discrete data, one in a fully observed model and the other
in a mixture model.  These examples illustrate how one might use the
posterior distribution of the partitions, given data and an assumed
generating process based on the distance dependent CRP.  We will focus
on models of discrete data and we will use the terminology of document
collections to describe these models.\footnote{While we focus on text,
  these models apply to any discrete data, such as genetic data, and,
  with modification, to non-discrete data as well.  That said,
  CRP-based methods have been extensively applied to text modeling and
  natural language
  processing~\citep{Teh:2007,Johnson:2007,Li:2007,Blei:2010}.}  Thus,
our observations are assumed to be collections of words from a fixed
vocabulary, organized into documents.

\paragraph{Language modeling.} In the language modeling application,
each document is associated with a distance dependent CRP, and its
tables are embellished with IID draws from a base distribution over
terms or words.  (The documents share the same base distribution.)  The
generative process of words in a document is as follows.  The data are
first placed at tables via customer assignments, and then assigned to
the word associated with their tables.  Subsets of the data exhibit a
partition structure by sharing the same table.

When using a traditional CRP, this is a formulation of a simple
Dirichlet-smoothed language model.  Alternatives to this model, such
as those using the Pitman-Yor process, have also been applied in this
setting~\citep{Teh:2006a,Goldwater:2006a}.  We consider a sequential
CRP, which assumes that a word is more likely to occur near itself in
a document.  Words are still considered contagious---seeing a word
once means we're likely to see it again---but the window of contagion
is mediated by the decay function.

More formally, given a decay function $f$, sequential distances $D$,
scaling parameter $\alpha$, and base distribution $G_0$ over discrete words,
$N$ words are drawn as follows,
\begin{packed_enumerate}
\item For each word $i \in \{1, \ldots, N\}$ draw assignment $c_i
  \sim \textrm{dist-CRP}(\alpha, f, D)$.
\item For each table, $k \in \{1, \ldots\}$, draw a word $w^* \sim
  G_0$.
\item For each word $i \in \{1, \ldots, N\}$, assign the word $w_i =
  w^*_{z(\bc)_i}$.
\end{packed_enumerate}
The notation $z(\bc)_i$ is the table assignment of the $i$th customer
in the table assignments induced by the complete collection of
customer assignments.

For each document, we observe a sequence of words $w_{1:N}$ from which
we can infer their seating assignments in the distance dependent CRP.
The partition structure of observations---that is, which words are the
same as other words---indicates either that they share the same table
in the seating arrangement, or that two tables share the same term
drawn from $G_0$.  We have not described the process sequentially, as
one would with a traditional CRP, in order to emphasize the three
stage process of the distance dependent CRP---first the customer
assignments and table parameters are drawn, and then the observations
are assigned to their corresponding parameter.  However, the
sequential distances $D$ guarantee that we can draw each word
successively.  This, in turn, means that we can easily construct a
predictive distribution of future words given previous words.  (See
\mysec{inference} below.)


\paragraph{Mixture modeling} \label{sec:mixture-modeling}

The second model we study is akin to the CRP mixture or (equivalently)
the DP mixture, but differs in that the mixture component for a data
point depends on the mixture component for nearby data.  Again, each
table is
endowed with a draw from a base distribution $G_0$, but here that draw is a
distribution over mixture component parameters.  In the document
setting, observations are documents (as opposed to individual words),
and $G_0$ is typically a Dirichlet distribution over
distributions of words~\citep{Teh:2007}.  The data are drawn as
follows:
\begin{packed_enumerate}
\item For each document $i \in [1,N]$ draw assignment $c_i \sim
  \textrm{dist-CRP}(\alpha, f, D)$.

\item For each table, $k \in \{1, \ldots\}$, draw a parameter
  $\theta_k^* \sim G_0$.

\item For each document $i \in [1,N]$, draw $w_i \sim F(\theta_{z(\bc)_i})$.
\end{packed_enumerate}
In \mysec{study}, we will study the sequential CRP in this setting,
choosing its structure so that contemporaneous documents are more
likely to be clustered together.  The distances $d_{ij}$ can be the
differences between indices in the ordering of the data, or lags
between external measurements of distance like date or time.  (Spatial
distances or distances based on other covariates can be used to define
more general mixtures, but we leave these settings for future work.)
Again, we have not defined the generative process sequentially but, as
long as $D$ respects the assumptions of a sequential CRP, an
equivalent sequential model is straightforward to define.


\paragraph{Relationship to dependent Dirichlet processes.}  More
generally, the distance dependent CRP mixture provides an alternative
to the dependent Dirichlet process (DDP) mixture as an infinite
clustering model that models dependencies between the latent component
assignments of the data~\citep{MacEachern:1999}.  The DDP has been
extended to sequential, spatial, and other kinds of
dependence~\citep{Griffin:2006,Duan:2006,Xue:2007}.  In all these
settings, statisticians have appealed to truncations of the
stick-breaking representation for approximate posterior inference,
citing the dependency between data as precluding the more efficient
techniques that integrate out the component parameters and
proportions.  In contrast, distance dependent CRP mixtures are
amenable to Gibbs sampling algorithms that integrate out these
variables (see \mysec{inference}).

An alternative to the DDP formalism is the Bayesian density regression
(BDR) model of~\cite{Dunson:2007}.  In BDR, each data point is
associated with a random measure and is drawn from a mixture of
per-data random measures where the mixture proportions are related to
the distance between data points.  Unlike the DDP, this model affords
a Gibbs sampler where the random measures can be integrated out.

However, it is still different in spirit from the distance dependent
CRP.  Data are drawn from distributions that are similar to
distributions of nearby data, and the particular values of nearby data
impose softer constraints than those in the distance dependent CRP.
As an extreme case, consider a random partition of the nodes of a
network, where distances are defined in terms of the number of hops
between nodes.  Further, suppose that there are several disconnected
components in this network, that is, pairs of nodes that are not
reachable from each other.  In the DDP model, these nodes are very
likely not to be partitioned in the same group.  In the ddCRP model,
however, it is impossible for them to be grouped together.

We emphasize that DDP mixtures (and BDR) and distance dependent CRP
mixtures are \textit{different} classes of models.  DDP mixtures are
Bayesian nonparametric models, interpretable as data drawn from a
random measure, while the distance dependent CRP mixtures generally
are not. DDP mixtures exhibit marginal invariance, while distance
dependent CRPs generally do not (see \mysec{marg}).  In their ability
to capture dependence, these two classes of models capture similar
assumptions, but the appropriate choice of model depends on the
modeling task at hand.



%% file: inference-alt.tex
\section{Posterior inference and prediction}
\label{sec:inference}


\newcommand{\hyper}{\eta} The central computational problem for distance dependent CRP
modeling is posterior inference, determining the conditional
distribution of the hidden variables given the observations.  This
posterior is used for exploratory analysis of the data and how it
clusters, and is needed to compute the predictive distribution of a
new data point given a set of observations.

Regardless of the likelihood model, the posterior will be intractable
to compute because the distance dependent CRP places a prior over a combinatorial
number of possible customer configurations.  In this section we
provide a general strategy for approximating the posterior using Monte
Carlo Markov chain (MCMC) sampling.  This strategy can be used in
either fully-observed or mixture settings, and can be used with
arbitrary distance functions.  (For example, in \mysec{study} we
illustrate this algorithm with both sequential distance functions and
graph-based distance functions and in both fully-observed and mixture
settings.)

In MCMC, we aim to construct a Markov chain whose stationary
distribution is the posterior of interest.  For distance dependent CRP models, the
state of the chain is defined by $c_i$, the customer assignments for
each data point.  We will also consider $z(\bc)$, which are the table
assignments that follow from the customer assignments (see
\myfig{tables}).  Let $\hyper = \{D, \alpha, f, G_0\}$ denote the set
of model hyperparameters.  It contains the distances $D$, the scaling
factor $\alpha$, the decay function $f$, and the base measure $G_0$.
Let $x$ denote the observations.

In Gibbs sampling, we iteratively draw from the conditional
distribution of each latent variable given the other latent variables
and observations.  (This defines an appropriate Markov chain, see
\cite{Neal:1993}.) In distance dependent CRP models, the Gibbs sampler iteratively
draws from
\begin{equation}
  \label{eq:gibbs}
  p(c_i^\textrm{(new)} \g \bc_{-i}, \bx, \hyper) \propto
  p(c_i^\textrm{(new)} \g D, \alpha) p(\bx \g z(\bc_{-i} \cup
  c_i^\textrm{(new)}),G_0).
\end{equation}
The first term is the distance dependent CRP prior from \myeq{dec-crp-prior}.

The second term is the likelihood of the observations under the
partition given by $z(\bc_{-i} \cup c_i^{(new)})$.  This can be
thought of as removing the current link from the $i$th customer and
then considering how each alternative new link affects the likelihood
of the observations.  Before examining this likelihood, we describe
how removing and then replacing a customer link affects the underlying
partition (i.e., table assignments).

\begin{figure}
  \begin{center}
    \includegraphics[width=0.85\textwidth]{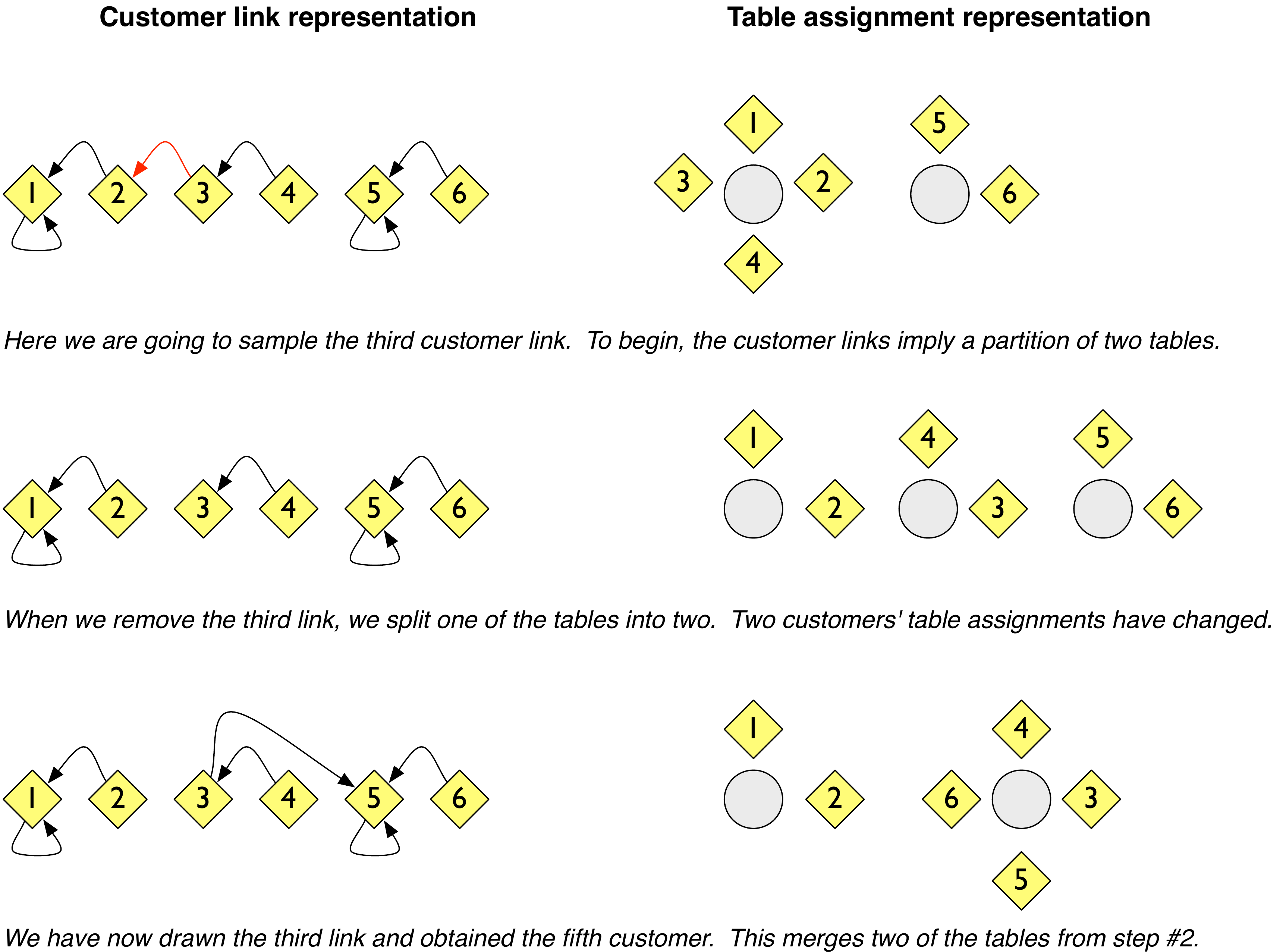}
  \end{center}
  \caption{\label{fig:inference-tutorial} An example of a single step
    of the Gibbs sampler.  Here we illustrate a scenario that
    highlights all the ways that the sampler can move: A table can be
    split when we remove the customer link before conditioning; and
    two tables can join when we resample that link.}
\end{figure}

To begin, consider the effect of removing a customer link.  What is
the difference between the partition $z(\bc)$ and $z(\bc_{-i})$?
There are two cases.

The first case is that a table splits.  This happens when $c_i$ is
the only connection between the $i$th data point and a particular
table.  Upon removing $c_i$, the customers at its table are split in
two: those customers pointing (directly or indirectly) to $i$ are at
one table; the other customers previously seated with $i$
are at a different table.  (See the change from the first to second rows of
\myfig{inference-tutorial}.)

The second case is that there is no change.  If the $i$th link is
not the only connection between customer $i$ and his table or if $c_i$ was a
self-link ($c_i=i$) then the tables remain the same.  In this case,
$z(\bc_{-i}) = z(\bc)$.

Now consider the effect of replacing the customer link.  What is the
difference between the partition $z(\bc_{-i})$ and $z(\bc_{-i} \cup
c_i^\textrm{(new)})$?  Again there are two cases.  The first case is
that $c_i^\textrm{(new)}$ joins two tables in $z(\bc_{-i})$.  Upon
adding $c_i^\textrm{(new)}$, the customers at its table become linked
to another set of customers.  (See the change from the second to
third rows of \myfig{inference-tutorial}.)

The second case, as above, is that there is no change.  This occurs if
$c_i^\textrm{(new)}$ points to a customer that is already at its table
under $z(\bc_{-i})$ or if $c_i^\textrm{(new)}$ is a self-link.

With the changed partition in hand, we now compute the likelihood
term.  We first compute the likelihood term for partition $z(\bc)$.
The likelihood factors into a product of terms, each of which is the
probability of the set of observations at each table.  Let $|z(\bc)|$
be the number of tables and $z^k(\bc)$ be the set of indices that are
assigned to table $k$.  The likelihood term is
\begin{equation}
  \label{eq:lhood}
  p(\bx \g z(\bc), G_0) =
  \prod_{k=1}^{|z(\bc)|} p(\bx_{z^k(\bc)} \g G_0).
\end{equation}

Because of this factorization, the Gibbs sampler need only compute
terms that correspond to changes in the partition.  Consider the
partition $z(\bc_{-i})$, which may have split a table, and the new
partition $z(\bc_{-i} \cup c^{\textrm{(new)}})$.  There are three
cases to consider.  First, $c_i$ might link to itself---there will be
no change to the likelihood function because a self-link cannot join
two tables.  Second, $c_i$ might link to another table but cause no
change in the partition.  Finally, $c_i$ might link to another table
and join two tables $k$ and $\ell$.  The Gibbs sampler for the
distance dependent CRP is thus
\begin{equation}
  p(c_i^\textrm{(new)}  \g \bc_{-i}, \bx, \hyper) \propto
  \left\{
    \begin{array}{ll}
      \alpha & \textrm{if } c_i^\textrm{(new)} \textrm{ is equal to } i. \\
      f(d_{ij}) & \textrm{if } c_i^\textrm{(new)}=j  \textrm{ does not join two
        tables.} \\
      f(d_{ij}) \large \frac{p(\bx_{z^k(\bc_{-i}) \cup z^\ell(\bc_{-i})} \g G_0)}
      {p(\bx_{z^k(\bc_{-i})} \g G_0) p(\bx_{z^\ell(\bc_{-i})} \g G_0)} &
      \textrm{if } c_i^\textrm{(new)}=j  \textrm{ joins tables } k \textrm{ and } \ell.
    \end{array}
  \right.
\end{equation}





The specific form of the terms in \myeq{lhood} depend on the model.
We first consider the fully observed case (i.e., ``language
modeling'').  Recall that the partition corresponds to words of the
same type, but that more than one table can contain identical types.
(For example, four tables could contain observations of the word
``peanut.''  But, observations of the word ``walnut'' cannot sit at
any of the peanut tables.)  Thus, the likelihood of the data is simply
the probability under $G_0$ of a representative from each table, e.g.,
the first customer, times a product of indicators to ensure that all
observations are equal,
\begin{equation}
  p(\bx_{z^k(\bc)} \g G_0) =  p(x_{z^k(\bc)_1} \g G_0) \textstyle
  \prod_{i \in z^k(\bc)} 1(x_i = x_{z^k(\bc)_1}),
\end{equation}
where $z^k(\bc)_1$ is the index of the first customer assigned to table $k$.

In the mixture model, we compute the marginal probability that the set
of observations from each table are drawn independently from the same
parameter, which itself is drawn from $G_0$.  Each term is
\begin{equation}
  p(\bx_{z^k(\bc)} \g G_0) =
  \int \textstyle \left(\prod_{i \in z^k(\bc)} p(x_i \g \theta)\right) p(\theta \g G_0) d\theta.
\end{equation}
Because this term marginalizes out the mixture component $\theta$, the
result is a collapsed sampler for the mixture model.
When $G_0$ and $p(x \g \theta)$ form a conjugate pair, the integral is
straightforward to compute.  In nonconjugate settings, an additional
layer of sampling is needed.

\paragraph{Prediction.}

In prediction, our goal is to compute the conditional probability
distribution of a new data point $x_{\rmnew}$ given the data set
$\bx$.  This computation relies on the posterior.  Recall that $D$ is
the set of distances between all the data points.  The predictive
distribution is
\begin{equation}
  p(x_{\rmnew} | \bx, D, G_0, \alpha) =
  \sum_{c_{\rmnew}} p(c_{\rmnew} \g D, \alpha)
  \textstyle \sum_{\bc}
  p(x_{\rmnew} | c_{\rmnew} , \bc, \bx, G_0)  p(\bc | \bx, D, \alpha, G_0).
\end{equation}

The outer summation is over the customer assignment of the new data
point; its prior probability only depends on the distance matrix $D$.
The inner summation is over the posterior customer assignments of the
data set; it determines the probability of the new data point
conditioned on the previous data and its partition.  In this
calculation, the difference between sequential distances and arbitrary
distances is important.

Consider sequential distances and suppose that $x_{\rmnew}$ is a
future data point.  In this case, the distribution of the data set
customer assignments $\bc$ does not depend on the new data point's
location in time.  The reason is that data points can only connect to
data points in the past.  Thus, the posterior $p(\bc \g \bx, D,
\alpha, G_0)$ is unchanged by the addition of the new data, and we can
use previously computed Gibbs samples to approximate it.

In other situations---nonsequential distances or sequential distances
where the new data occurs somewhere in the middle of the
sequence---the discovery of the new data point changes the posterior
$p(\bc \g \bx, D, \alpha, G_0)$.  The reason is that the knowledge of
where the new data is relative to the others (i.e., the information in
$D$) changes the prior over customer assignments and thus changes the
posterior as well.  This new information requires rerunning the Gibbs
sampler to account for the new data point.  Finally, note that the
special case where we know the new data's location in advance (without
knowing its value) does not require rerunning the Gibbs sampler.

%% file: marg-invariance.tex
\section{Marginal invariance}
\label{sec:marg}

In \mysec{model} we discussed the property of \textit{marginal
invariance}, where removing a customer leaves the partition
distribution over the remaining customers unchanged.  When a model has
this property, unobserved data may simply be ignored.  We mentioned
that the traditional CRP is marginally invariant, while the distance
dependent CRP does not necessarily have this property.

In fact, the traditional CRP is the {\it only} distance dependent CRP
that is marginally invariant.\footnote{One can also create a marginally
invariant distance dependent CRP by combining several
independent copies of the traditional CRP.  Details are discussed in
the appendix.}  The details of this characterization are given in the
appendix.  This characterization of marginally invariant CRPs contrasts
the distance dependent CRP with the alternative priors over partitions
induced by random measures, such as the Dirichlet process. 

In addition to the Dirichlet process, random-measure models include the
dependent Dirichlet process~\citep{MacEachern:1999} and the order-based
dependent Dirichlet process~\citep{Griffin:2006}.  
These models suppose that data from a given covariate were drawn
independently from a fixed latent sampling probability measure.
These models then suppose that these sampling measures were drawn from
some parent probability measure.  Dependence between the randomly drawn
sampling measures is achieved through this parent probability measure. 


We formally define a random-measure model as follows.  Let $\X$ and
$\Y$ be the sets in which covariates and observations take their
values, let $x_{1:N}\subset\X$, $y_{1:N}\subset\Y$ be the set of
observed covariates and their corresponding sampled values, and let
$M(\Y)$ be the space of probability measures on $\Y$.  A
random-measure model is any probability distribution on the samples
$y_{1:N}$ induced by a probability measure $G$ on the space
$M(\Y)^{\X}$.  This random-measure model may be written
\begin{equation}
\label{eq:MeasureOnMeasures}
y_n \mid x_n \sim \Pmeasure_{x_n},\qquad
\left(\Pmeasure_x\right)_{x\in\X} \sim G,
\end{equation}
where the $y_n$ are conditionally independent of each other given $(\Pmeasure_x)_{x\in\X}$.
Such models implicitly induce a distribution on partitions of the data
by taking all points $n$ whose sampled values $y_n$ are equal to be in
the same cluster.

In such random-measure models, the (prior) distribution on $y_{-n}$
does not depend on $x_{n}$, and so such models are
marginally invariant, regardless of the points $x_{1:n}$ and the
distances between them.  From this observation, and the lack of marginal
invariance of the distance dependent CRP, it follows that the distributions
on partitions induced by random-measure models are different from 
the distance dependent CRP.  The only distribution that is both a distance
dependent CRP, and is also induced by a random-measure model, is the traditional
CRP.

Thus, distance dependent CRPs are generally not marginally invariant, and so
are appropriate for modeling situations that naturally depart from marginal
invariance.  This distinguishes priors obtained with distance dependent CRPs
from those obtained from random-measure models, which are appropriate when
marginal invariance is a reasonable assumption.

%% file: study.tex
\section{Empirical study}
\label{sec:study}

We studied the distance dependent CRP in the language modeling and
mixture settings on four text data sets.  We explored both time
dependence, where the sequential ordering of the data is respected via
the decay function and distance measurements, and network dependence,
where the data are connected in a graph.  We show below that the
distance dependent CRP gives better fits to text data in both the
fully-observed and mixture modeling settings.\footnote{Our R
  implementation of Gibbs sampling for ddCRP models is available at\\
  \url{http://www.cs.princeton.edu/~blei/downloads/ddcrp.tgz}}

Further, we compared the traditional Gibbs sampler for DP mixtures to
the Gibbs sampler for the distance dependent CRP formulation of DP
mixtures.  We found that the sampler based on customer assignments
mixes faster than the traditional sampler.

\subsection{Language modeling}

\begin{figure*}[t]
  \begin{center}
    \includegraphics[width=\textwidth]{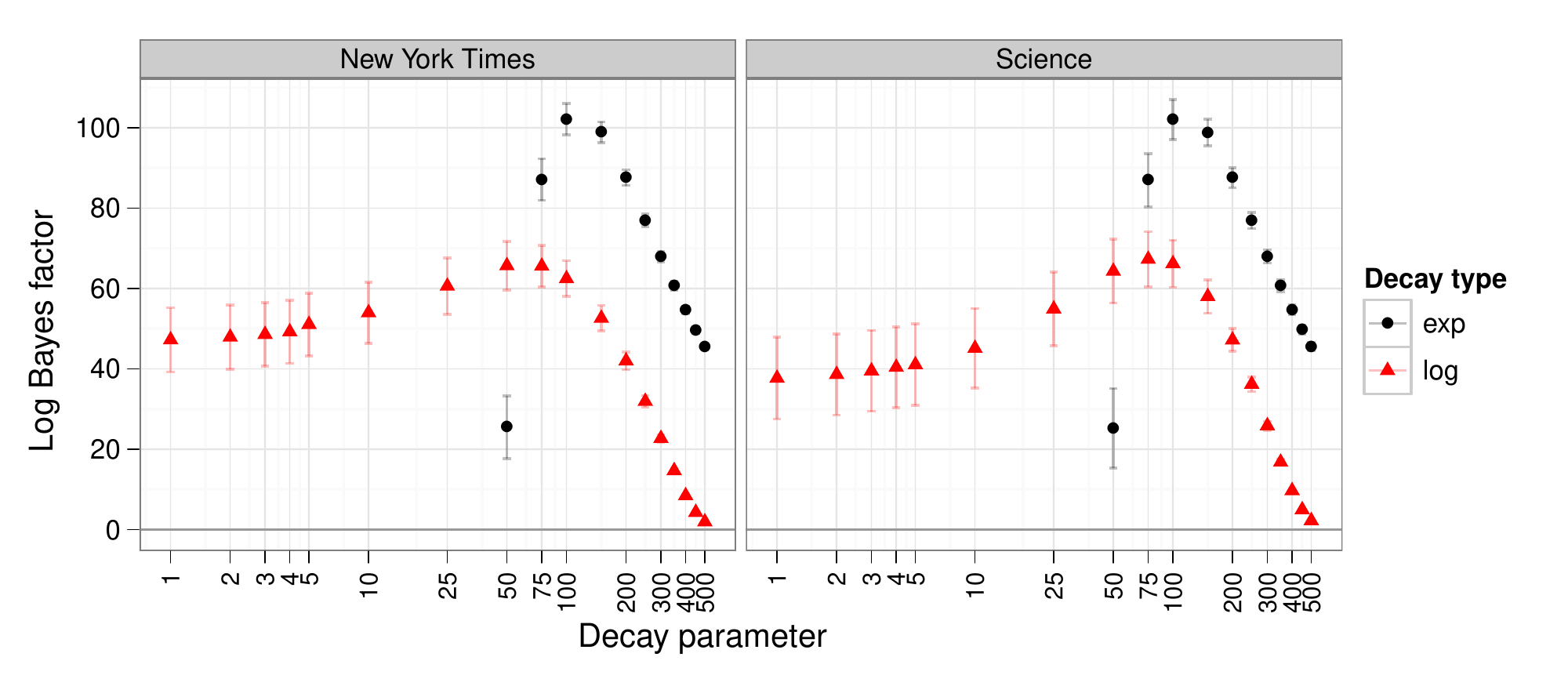}
  \end{center}
  \caption{ \label{fig:bayesfactors} Bayes factors of the distance
    dependent CRP versus the traditional CRP on documents from
    \textit{Science} and the \textit{New York Times}.  The black line
    at $0$ denotes an equal fit between the traditional CRP and
    distance dependent CRP, while positive values denote a better fit
    for the distance dependent CRP. Also illustrated are standard
    errors across documents.  }
\end{figure*}

We evaluated the fully-observed distance dependent CRP models on two
data sets: a collection of 100 OCR'ed documents from the journal
\textit{Science} and a collection of 100 world news articles from the
\textit{New York Times}.  We modeled each document independently.  We
assess sampler convergence visually, examining the autocorrelation
plots of the log likelihood of the state of the
chain~\citep{Robert:2004}.


We compare models by estimating the Bayes factor, the ratio of the
probability under the distance dependent CRP to the probability under
the traditional CRP~\citep{Kass:1995}.  For a decay function $f$, this
Bayes factor is
\begin{equation}
  BF_{f,\alpha} =
  p(w_{1:N} \g \textrm{dist-CRP}_{f, \alpha})/p(w_{1:N} \g
  \textrm{CRP}_{\alpha}).
\end{equation}
A value greater than one indicates an improvement of the distance
dependent CRP over the traditional CRP.  Following~\cite{Geyer:1992},
we estimate this ratio with a Monte Carlo estimate from posterior
samples.

\myfig{bayesfactors} illustrates the average log Bayes factors across
documents for various settings of the exponential and logistic decay
functions.  The logistic decay function always provides a better model
than the traditional CRP; the exponential decay function provides a
better model at certain settings of its parameter.  (These curves are
for the hierarchical setting with the base distribution over terms
$G_0$ unobserved; the shapes of the curves are similar in the
non-hierarchical settings.)


\subsection{Mixture modeling}
\label{sec:mixture-study}

\begin{figure*}[tb]
  \begin{center}
      \includegraphics[width=\textwidth]{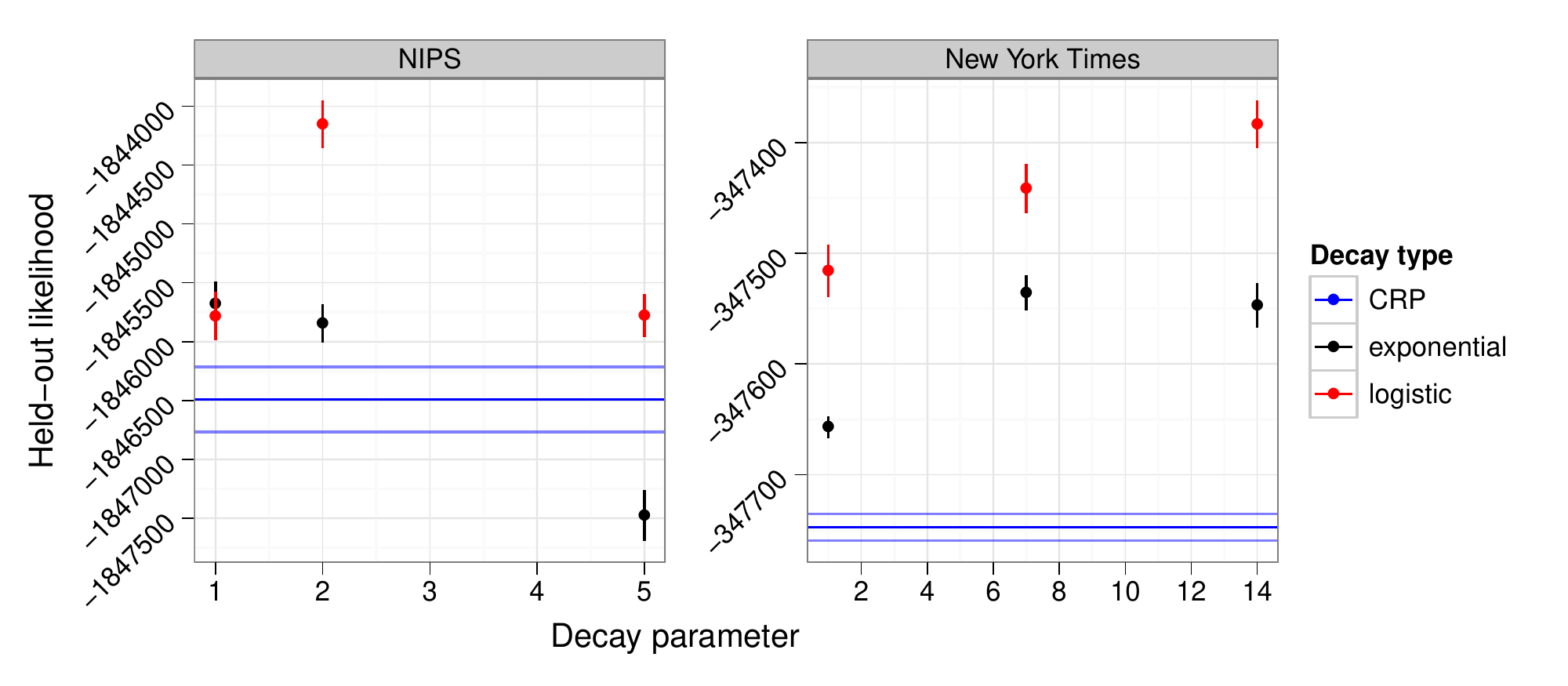}
    \end{center}
    \caption{Predictive held-out log likelihood for the last year of
      NIPS and last three days of the \textit{New York Times} corpus.
      Error bars denote standard errors across MCMC samples.  On the
      NIPS data, the distance dependent CRP outperforms the
      traditional CRP for the logistic decay with a decay parameter of $2$ years.
      On the \textit{New York Times} data, the distance dependent CRP
      outperforms the traditional CRP in almost all settings
      tested. \label{fig:mixt}}
\end{figure*}

We examined the distance dependent CRP mixture on two text corpora.
We analyzed one month of the \textit{New York Times} (NYT) time-stamped
by day, containing 2,777 articles, 3,842 unique terms and 530K
observed words.  We also analyzed 12 years of NIPS papers time-stamped
by year, containing 1,740 papers, 5,146 unique terms, and 1.6M
observed words. Distances $D$ were differences between time-stamps.

In both corpora we removed the last 250 articles as held out data.  In
the NYT data, this amounts to three days of news; in the NIPS data,
this amounts to papers from the 11th and 12th year.  (We retain the time
stamps of the held-out articles because the predictive likelihood of an
article's contents depends on its time stamp, as well as the time stamps of
earlier articles.)
We evaluate the models by
estimating the predictive likelihood of the held out data.  The
results are in \myfig{mixt}.  On the NYT corpus, the distance
dependent CRPs definitively outperform the traditional CRP.  A
logistic decay with a window of 14 days performs best.  On the NIPS
corpus, the logistic decay function with a decay parameter of 2 years
outperforms the traditional CRP.  In general, these results show that
non-exchangeable models given by the distance dependent CRP mixture
provide a better fit than the exchangeable CRP mixture.

\subsection{Modeling networked data}

The previous two examples have considered data analysis settings with
a sequential distance function.  However, the distance dependent CRP
is a more general modeling tool.  Here, we demonstrate its flexibility
by analyzing a set of \textit{networked documents} with a distance
dependent CRP mixture model.  Networked data induces an entirely
different distance function, where any data point may link to an
arbitrary set of other data.  We emphasize that we can use the same
Gibbs sampling algorithms for both the sequential and networked
settings.

Specifically, we analyzed the CORA data set, a collection of Computer
Science abstracts that are connected if one paper cites the
other~\citep{McCallum:2000}.  One natural distance function is the
number of connections between data (and $\infty$ if two data points
are not reachable from each other).  We use the window decay function
with parameter $1$, enforcing that a customer can only link to itself
or to another customer that refers to an immediately connected
document.  We treat the graph as undirected.

\begin{figure}
  \begin{center}
    \includegraphics[width=\textwidth]{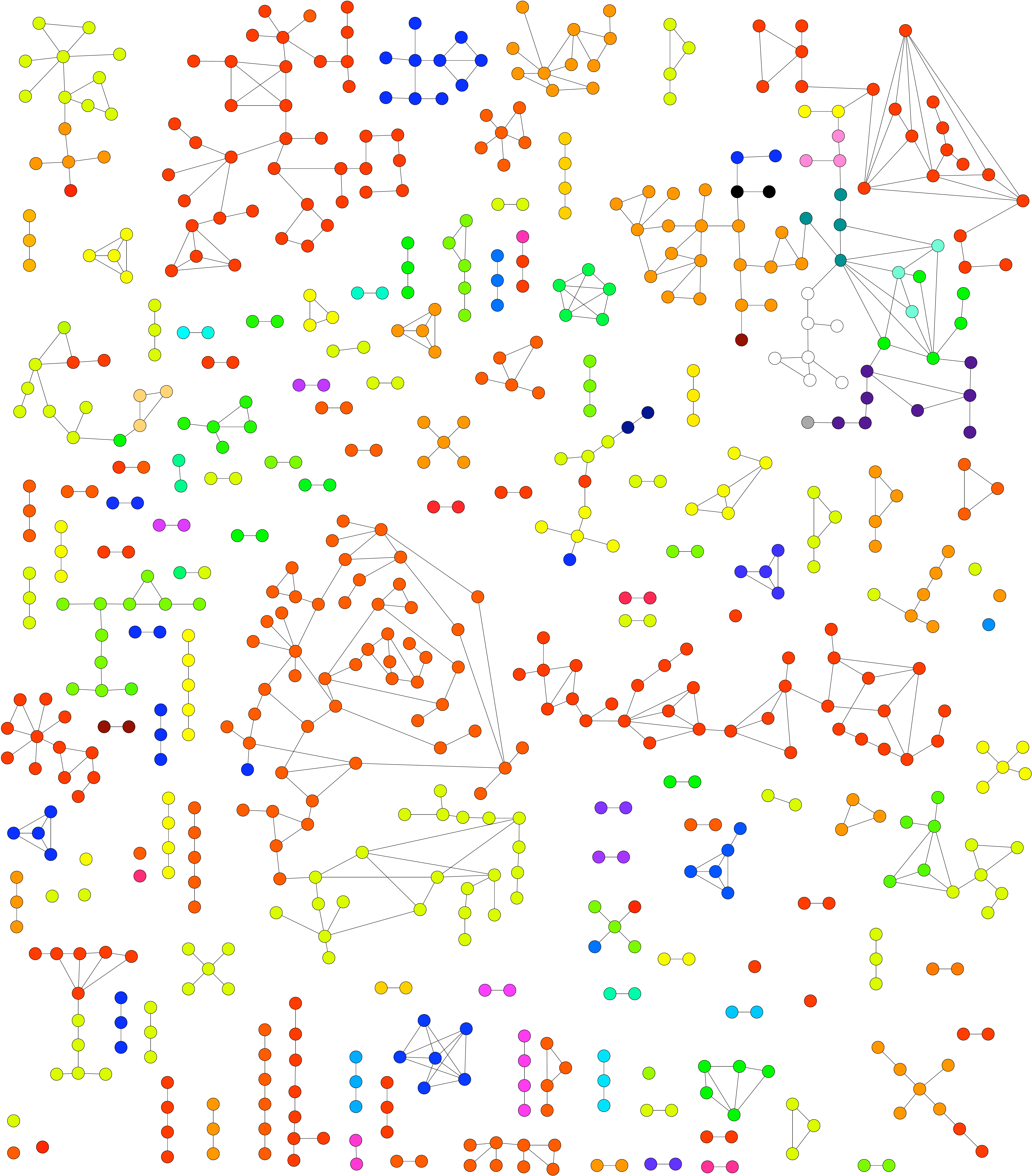}
  \end{center}
  \caption{\label{fig:cora} The MAP clustering of a subset of CORA.
    Each node is an abstract in the collection and each link
    represents a citation.  
    Colors are repeated across connected components -- no two data points from disconnected components in the graph
    can be assigned to the same cluster. Within each connected component, colors are not repeated, and nodes with
    the same color are assigned to the same cluster.}
\end{figure}

\myfig{cora} shows a subset of the MAP estimate of the clustering
under these assumptions.  Note that the clusters form connected groups
of documents, though several clusters are possible within a large
connected group.  Traditional CRP clustering does not lean towards
such solutions.  Overall, the distance dependent CRP provides a better model.  The log
Bayes factor is 13,062, strongly in favor of the distance dependent CRP,
although we emphasize that much of this improvement may occur simply
because the distance dependent CRP avoids clustering abstracts from unconnected components of the network.
Further analysis is needed to understand the abilities of the distance dependent CRP 
beyond those of simpler network-aware clustering schemes.

We emphasize that this analysis is meant to be a proof of concept to
demonstrate the flexibility of distance dependent CRP mixtures.  Many
modeling choices can be explored, including longer windows in the
decay function and treating the graph as a directed graph.  A similar
modeling set-up could be used to analyze spatial data, where distances
are natural to compute, or images (e.g., for image segmentation),
where distances might be the Manhattan distance between pixels.

\subsection{Comparison to the traditional Gibbs sampler}
\label{sec:crp-comparison}

The distance dependent CRP can express a number of flexible models.
However, as we describe in \mysec{model}, it can also re-express the
traditional CRP.  In the mixture model setting, the Gibbs sampler of
\mysec{inference} thus provides an alternative algorithm for
approximate posterior inference in DP mixtures.  We compare this Gibbs
sampler to the widely used collapsed Gibbs sampler for DP mixtures,
i.e., Algorithm 3 from~\cite{Neal:2000}, which is applicable when the
base measure $G_0$ is conjugate to the data generating distribution.

The Gibbs sampler for the distance dependent CRP iteratively samples
the customer assignment of each data point, while the
collapsed Gibbs sampler iteratively samples the cluster assignment of
each data point.  The practical difference between the two algorithms
is that the distance dependent CRP based sampler can change several
customers' cluster assignments via a single customer assignment.  This
allows for larger moves in the state space of the posterior and, we
will see below, faster mixing of the sampler.

Moreover, the computational complexity of the two samplers is the
same.  Both require computing the change in likelihood of adding or
removing either a set of points (in the distance dependent CRP case)
or a single point (in the traditional CRP case) to each cluster.
Whether adding or removing one or a set of points, this amounts to
computing a ratio of normalizing constants for each cluster, and this
is where the bulk of the computation of each sampler
lies.\footnote{In some settings, removing a single point---as
  is done in \cite{Neal:2000}---allows faster computation of each sampler
  iteration.  This is true, for example, if the observations are single
  words (as opposed to a document of words) or single draws from a Gaussian.
  Although each iteration may be faster with the traditional sampler, that
  sampler may spend many more iterations stuck in local optima.}

\begin{figure}[tb]
  \begin{tabular}{cc}
    \includegraphics[width=0.45\textwidth]{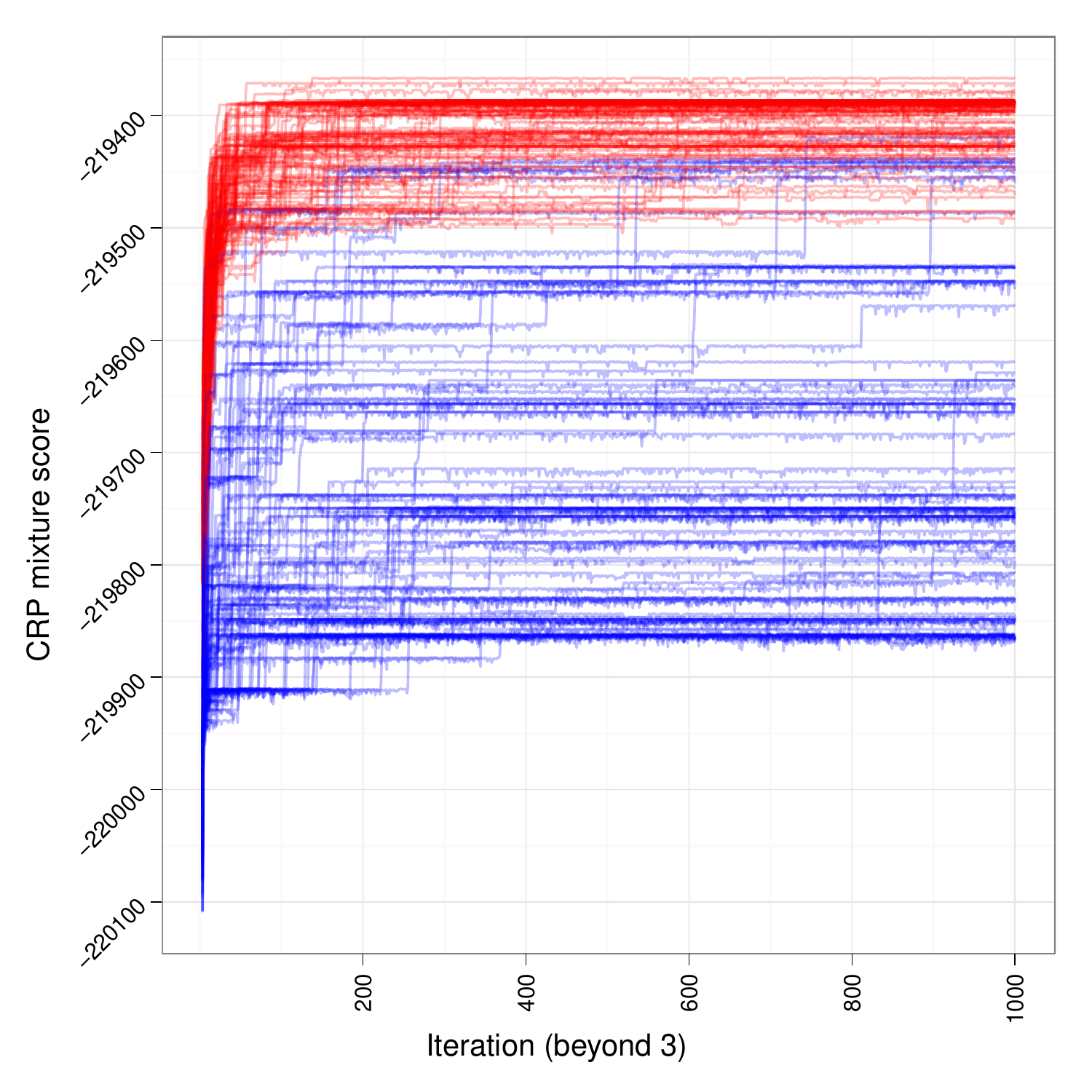}
    \includegraphics[width=0.45\textwidth]{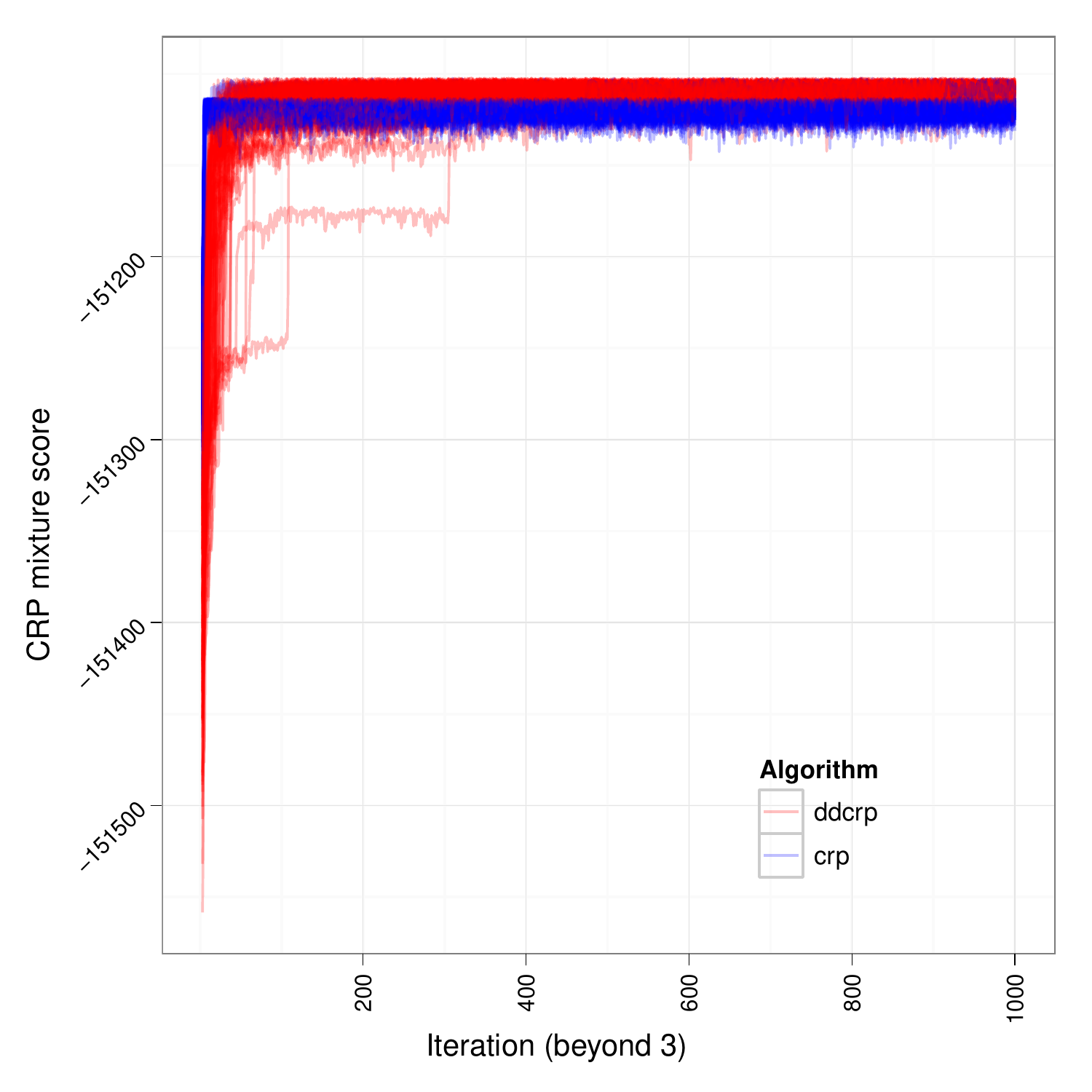}
  \end{tabular}
  \caption{\label{fig:crp-comparison} Each panel illustrates 100 Gibbs
    runs using Algorithm 3 of ~\citep{Neal:2000} (CRP, in blue) and the
    sampler from \mysec{inference} with the identity decay function
    (distance dependent CRP, in red).  Both samplers have the same limiting distribution
    because the distance dependent CRP with identity decay is the
    traditional CRP.   We plot the log probability of the CRP representation
    (i.e., the divergence) as a function of its iteration.  The left panel
    shows the \textit{Science} corpus, and the right panel shows the
    \textit{New York Times} corpus.  Higher values indicate that the chain
    has found a better local mode of the posterior.  In these examples, the 
    distance dependent CRP Gibbs sampler mixes faster.}
\end{figure}

To compare the samplers, we analyzed documents from the
\textit{Science} and \textit{New York Times} collections under a CRP
mixture with scaling parameter equal to one and uniform Dirichlet base
measure.  \myfig{crp-comparison} illustrates the log probability of
the state of the traditional CRP Gibbs sampler as a function of Gibbs
sampler iteration.  The log probability of the state is proportional
to the posterior; a higher value indicates a state with higher
posterior likelihood.  These numbers are comparable because the
models, and thus the normalizing constant, are the same for both the
traditional representation and customer based CRP.  Iterations 3--1000
are plotted, where each sampler is started at the same (random) state.
The traditional Gibbs sampler is much more prone to stagnation at local
optima, particularly for the \textit{Science} corpus.


%% file: discussion.tex
\section{Discussion}

We have developed the distance dependent Chinese restaurant process, a
distribution over partitions that accommodates a flexible and
non-exchangeable seating assignment distribution.  The distance
dependent CRP hinges on the customer assignment representation.  We
derived a general-purpose Gibbs sampler based on this representation,
and examined sequential models of text.

The distance dependent CRP opens the door to a number of further
developments in infinite clustering models.  We plan to explore
spatial dependence in models of natural images, and multi-level models
akin to the hierarchical Dirichlet process~\citep{Teh:2007}.
Moreover, the simplicity and fixed dimensionality of the corresponding
Gibbs sampler suggests that a variational method is worth exploring as
an alternative deterministic form of approximate inference.

\section*{Acknowledgments}

David M. Blei is supported by ONR 175-6343, NSF CAREER 0745520, AFOSR
09NL202, the Alfred P. Sloan foundation, and a grant from Google.
Peter I. Frazier is supported by AFOSR YIP FA9550-11-1-0083.  Both
authors thank the three anonymous reviewers for their insightful
comments and suggestions.



%% file: proofs-alt.tex
\section{A formal characterization of marginal invariance}
\label{sec:formal-marg}

In this section, we formally characterize the class of distance
dependent CRPs that are marginally invariant.  This family is a very
small subset of the entire set of distance dependent CRPs, containing
only the traditional CRP and variants constructed from independent
copies of it.  This characterization is used in \mysec{marg} to
contrast the distance dependent CRP with random-measure models.

Throughout this section, we assume that the decay function satisfies
a relaxed version of the triangle inequality, which uses the notation $\bar{d}_{ij}=\min(d_{ij},d_{ji})$.
We assume:
if $\bar{d}_{ij}=0$ and $\bar{d}_{jk}=0$ then $\bar{d}_{ik}=0$; and if
$\bar{d}_{ij}<\infty$ and $\bar{d}_{jk}<\infty$ then $\bar{d}_{ik}<\infty$.

\subsection{Sequential Distances}

\newcommand{\Jfunc}{J}
\newcommand{\Jset}{\mathcal{J}}
\newcommand{\Kfunc}{k}
\newcommand{\Ksz}{K}

We first consider sequential distances.  We begin with the following proposition, which shows that a very restricted class of distance dependent CRPs may also be constructed by collections of independent CRPs. 

\begin{proposition}
  \label{p:sequential-crp}
  Fix a set of sequential distances between each of $n$ customers, a real number $a>0$, and a set $A\in\{\emptyset, \{0\}, \R\}$.  
  Then there is a (non-random) partition $B_1,\ldots,B_K$ of $\{1,\ldots,n\}$ for which two distinct customers $i$ and $j$ are in the same set $B_k$ iff $\bar{d}_{ij} \in A$.
  For each $k=1,\ldots,K$, let there be an independent CRP with concentration parameter $\alpha/a$, and let customers within $B_k$ be clustered among themselves according to this CRP.

  Then, the probability distribution on clusters induced by this construction is identical to the distance dependent CRP 
  with decay function $f(d) = a \ind{d\in A}$.  Furthermore, this probability distribution is marginally invariant.
\end{proposition}
\begin{proof}
We begin by constructing a partition $B_1,\ldots,B_K$ with the stated property.
Let $\Jfunc(i) = \min\{ j : \text{$j=i$ or $\bar{d}_{ij}\in A$}\}$, and let $\Jset = \{\Jfunc(i) : i=1,\ldots,n\}$ be the set of unique values taken by $\Jfunc$.
Each customer $i$ will be placed in the set containing customer $J(i)$.
Assign to each value $j\in\Jset$ a unique integer $\Kfunc(j)$ between $1$ and $|\Jset|$.
For each $j\in\Jset$, let $B_{\Kfunc(j)} = \{ i : \Jfunc(i) = j \} = \{ i : \text{$i=j$ or $\bar{d}_{ij} \in A$} \}$.
Each customer $i$ is in exactly one set, $B_{\Kfunc(\Jfunc(i))}$, and so $B_1,\ldots,B_{|\Jset|}$ is a partition of $\{1,\ldots,n\}$.

To show that $i\ne i'$ are both in $B_k$ iff $\bar{d}_{ii'} \in A$, we consider two possibilties.  
If $A=\emptyset$, then $\Jfunc(i) = i$ and each $B_k$ contains only a single point.
If $A=\{0\}$ or $A=\R$, then it follows from the relaxed triangle inequality assumed at the beginning of Appendix~\ref{sec:formal-marg}.

With this partition $B_1,\ldots,B_K$, the probability of linkage under the
distance dependent CRP with decay function $f(d) = a \ind{d\in A}$ may be written
\begin{equation*}
p(c_i = j) \propto
\begin{cases}
\alpha & \text{if $i=j$,} \\
a & \text{if $j<i$ and $j\in B_{\Kfunc(i)}$,} \\
0 & \text{if $j>i$ or $j\notin B_{\Kfunc(i)}$.}
\end{cases}
\end{equation*}

By noting that linkages between customers from different sets $B_k$ 
occur with probability $0$, we see that this is the same probability
distribution produced by taking $K$ independent distance dependent CRPs,
where the $k$th distance dependent CRP governs linkages between
customers in $B_k$ using
\begin{equation*}
p(c_i = j) \propto
\begin{cases}
\alpha & \text{if $i=j$,} \\
a & \text{if $j<i$,} \\
0 & \text{if $j>i$,}
\end{cases}
\end{equation*}
for $i,j\in B_k$.

Finally, dividing the unnormalized probabilities by $a$, we rewrite the
linkage probabilities for the $k$th distance dependent CRP as
\begin{equation*}
p(c_i = j) \propto
\begin{cases}
\alpha/a & \text{if $i=j$,} \\
1 & \text{if $j<i$,} \\
0 & \text{if $j>i$,}
\end{cases}
\end{equation*}
for $i,j\in B_k$.  This is identical to the distribution of the traditional CRP with concentration parameter $\alpha/a$.

This shows that the distance dependent CRP with decay function
$f(d)=a\ind{d\in A}$ induces the same probability distribution on
clusters as the one produced by a collection of $K$ independent
traditional CRPs, each with concentration parameter $\alpha/a$, where
the $k$th traditional CRP governs the clusters of customers within
$B_k$.

The marginal invariance of this distribution follows from the marginal invariance of each traditional CRP, and their independence from one another.
\end{proof}

The probability distribution described in this proposition separates
customers into groups $B_1,\ldots,B_K$ based on whether inter-customer
distances fall within the set $A$, and then governs clustering within
each group independently using a traditional CRP.  Clustering across
groups does not occur.

We consider what this means for specific choices of $A$.  If $A=\{0\}$,
then each group contains those customers whose distance from one another
is $0$.  This group is well-defined because of the assumption that
$d_{ij}=0$ and $d_{jk}=0$ implies $d_{ik}=0$. If $A=\R$, then each group
contains those customers whose distance from one another is finite.
Similarly to the $A=\{0\}$ case, this group is well-defined because of
the assumption that $d_{ij}<\infty$ and $d_{jk}<\infty$ implies
$d_{ik}<\infty$.  If $A=\emptyset$, then each group contains only a
single customer.  In this case, each customer will be in his own cluster.

Since the resulting construction is marginally invariant,
Proposition~\ref{p:sequential-crp} provides a sufficient condition for
marginal invariance.  The following proposition shows that this
condition is necessary as well.

\begin{proposition}
  \label{p:sequential-marginal-invariance}
  If the distance dependent CRP for a given decay function $f$ is
  marginally invariant over all sets of sequential distances then $f$ is
  of the form $f(d) = a \ind{d\in A}$ for some $a>0$ and $A$ equal to
  either $\emptyset$, $\{0\}$, or $\R$.
\end{proposition}
\begin{proof}
Consider a setting with $3$ customers, in which customer $2$ may either be
absent, or present with his seating assignment marginalized out.  Fix a
non-increasing decay function $f$ with $f(\infty)=0$ and suppose that the
distances are sequential, so $d_{13}=d_{23}=d_{12}=\infty$.  Suppose that the
distance dependent CRP resulting from this $f$ and any collection of sequential
distances is marginally invariant.  Then the probability that customers $1$
and $3$ share a table must be the same whether customer $2$ is absent or
present.

If customer $2$ is absent,
\begin{equation}
\label{eq:2absent}
\Prob{\text{$1$ and $3$ sit at same table} \mid \text{$2$ absent}}
= \frac{f(d_{31})}{f(d_{31}) + \alpha}.
\end{equation}

If customer $2$ is present, customers $1$ and $3$ may sit at the same
table in two different ways: $3$ sits with $1$ directly ($c_3=1$); or $3$
sits with $2$, and $2$ sits with $1$ ($c_3=2$ and $c_2=1$).  Thus,
\begin{multline}
\label{eq:2present}
\Prob{\text{$1$ and $3$ sit at same table} \mid \text{$2$ present}}\\
= \frac{f(d_{31})}{f(d_{31}) + f(d_{32}) + \alpha}
+ \left(\frac{f(d_{32})}{f(d_{31}) + f(d_{32}) + \alpha}\right)
  \left(\frac{f(d_{21})}{f(d_{21}) + \alpha}\right).
\end{multline}

For the distance dependent CRP to be marginally invariant,
\myeq{2absent} and \myeq{2present} must be identical.  Writing
\myeq{2absent} on the left side and \myeq{2present} on the right, we
have
\begin{equation}
\frac{f(d_{31})}{f(d_{31}) + \alpha}
= \frac{f(d_{31})}{f(d_{31}) + f(d_{32}) + \alpha}
+ \left(\frac{f(d_{32})}{f(d_{31}) + f(d_{32}) + \alpha}\right)
  \left(\frac{f(d_{21})}{f(d_{21}) + \alpha}\right).
\label{eq:equality}
\end{equation}

We now consider two different possibilities for the distances $d_{32}$ and $d_{21}$, always keeping $d_{31}=d_{21}+d_{32}$.


First, suppose $d_{21} = 0$ and $d_{32} = d_{31} = d$ for some $d\ge 0$.
By multiplying \myeq{equality} through by 
$\left(2f(d) + \alpha\right)\left(f(0)+\alpha\right)\left(f(d)+\alpha\right)$ 
and rearranging terms, we obtain
\begin{equation*}
  0 = \alpha f(d)\left( f(0) - f(d) \right).
\end{equation*}
Thus, either $f(d)=0$ or $f(d)=f(0)$.  Since this is true for each
$d\ge0$ and $f$ is nonincreasing, $f=a\ind{d\in A}$ with $a\ge0$ and either $A=\emptyset$, $A=\R$, $A=[0,b]$, or $A=[0,b)$ with $b\in[0,\infty)$.  Because $A=\emptyset$ is among the choices, we may assume $a>0$ without loss of generality.  We now show that if $A=[0,b]$ or $A=[0,b)$, then we must have $b=0$ and $A$ is of the form claimed by the proposition.

Suppose for contradiction that $A=[0,b]$ or $A=[0,b)$ with $b>0$.  
Consider distances given by $d_{32}=d_{21}=d=b-\epsilon$ with $\epsilon\in(0,b/2)$.  
By multiplying
\myeq{2present} through by 
\begin{equation*}
  \left(f(2d) + f(d) + \alpha\right)\left(f(d)+\alpha\right)\left(f(2d)+\alpha\right)
\end{equation*}
and rearranging terms, we obtain
\begin{equation*}
  0 = \alpha f(d)\left( f(d) - f(2d) \right).
\end{equation*}
Since  $f(d)=a>0$, we must have $f(2d)=f(d)>0$.
But, $2d = 2(b - \epsilon) > b$ implies together with $f(2d) = a \ind{2d\in A}$
that $f(2d)=0$, which is a contradiction.
\end{proof}

These two propositions are combined in the following corollary, which
states that the class of decay functions considered in
Propositions~\ref{p:sequential-crp}
and~\ref{p:sequential-marginal-invariance} is both necessary and
sufficient for marginal invariance.

\begin{corollary}
  \label{c:sequential-not-marg}
  Fix a particular decay function $f$.
  The distance dependent CRP resulting from this decay function is marginally invariant over all sequential distances if and only if $f$ is of the form $f(d) = a\ind{d\in A}$ for some $a>0$ and some $A\in\{\emptyset, \{0\}, \R\}$.
\end{corollary}
\begin{proof}
  Sufficiency for marginal invariance is shown by Proposition~\ref{p:sequential-crp}.  Necessity is shown by Proposition~\ref{p:sequential-marginal-invariance}.
\end{proof}

Although Corollary~\ref{c:sequential-not-marg} allows any choice of $a>0$ in the decay function $f(d)=a\ind{d\in A}$, the distribution of the distance dependent CRP with a particular $f$ and $\alpha$ remains unchanged if both $f$ and $\alpha$ are multiplied by a constant factor (see \myeq{dec-crp-prior}).  Thus, the distance dependent CRP defined by $f(d)=a\ind{d\in A}$ and concentration parameter $\alpha$ is identical to the one defined by $f(d)=\ind{d\in A}$ and concentration parameter $\alpha/a$.  In this sense, we can restrict the choice of $a$ in Corollary~\ref{c:sequential-not-marg} (and also Propositions~\ref{p:sequential-crp} and~\ref{p:sequential-marginal-invariance})
to $a=1$ without loss of generality.


\subsection{General Distances}

We now consider all sets of distances, including non-sequential
distances.  The class of distance dependent CRPs that are marginally
invariant over this larger class of distances is even more restricted
than in the sequential case.  We have the following proposition
providing a necessary condition for marginal invariance.

\begin{proposition}
  \label{p:general-marginal-invariance}
  If the distance dependent CRP for a given decay function $f$ is
  marginally invariant over all sets of distances, both sequential and
  non-sequential, then $f$ is identically $0$.
\end{proposition}
\begin{proof}
  From Proposition~\ref{p:sequential-marginal-invariance}, we have that any decay function that is marginally invariant under all sequential distances must be of the form $f(d) = a \ind{d\in A}$, where $a>0$ and $A\in\{\emptyset,\{0\},\R\}$.  We now show that if the decay function is marginally invariant under {\it all} sets of distances (not just those that are sequential), then $f(0)=0$.  The only decay function of the form $f(d) = a \ind{d\in A}$ that satisfies $f(0)=0$ is the one that is identically $0$, and so this will show our result.

  To show $f(0) = 0$, suppose that we have $n+1$ customers, all of whom are a distance $0$ away from one another, so $d_{ij}=0$ for $i,j=1,\ldots,n+1$.  Under our assumption of marginal invariance, the probability that the first $n$ customers sit at separate tables should be invariant to the absence or presence of customer $n+1$. 

  When customer $n+1$ is absent, the only way in which the first $n$
  customers may sit at separate tables is for each to link to himself.
  Let $p_n = \alpha / (\alpha + (n-1)f(0))$ denote the probability of a
  given customer linking to himself when customer $n+1$ is absent.
  Then 
  \begin{equation}
    \label{eq:general-distance-proof-absent}
    \Prob{\text{$1,\ldots,n$ sit separately} \mid \text{$n+1$ absent}}
    = (p_n)^n.
  \end{equation}

  We now consider the case when customer $n+1$ is present.  Let $p_{n+1}
  = \alpha / (\alpha+nf(0))$ be the probability of a given customer
  linking to himself, and let $q_{n+1} = f(0) / (\alpha+nf(0))$ be the
  probability of a given customer linking to some other given customer.
  The first $n$ customers may each sit at separate tables in two
  different ways.  First, each may link to himself, which occurs
  with probability $(p_{n+1})^n$.  Second, all but one of these first
  $n$ customers may link to himself, with the remaining customer
  linking to customer $n+1$, and customer $n+1$ linking either to
  himself or to the customer that linked to him.  This
  occurs with probability $n (p_{n+1})^{n-1} q_{n+1} (p_{n+1} +
  q_{n+1})$.
  Thus, the total probability that the first $n$ customers sit at separate tables is 
  \begin{equation}
    \label{eq:general-distance-proof-present}
    \begin{split}
    \Prob{\text{$1,\ldots,n$ sit separately} \mid \text{$n+1$ present}}
    = (p_{n+1})^n + n (p_{n+1})^{n-1} q_{n+1} (p_{n+1} + q_{n+1}).
    \end{split}
  \end{equation}

  Under our assumption of marginal invariance,
  \myeq{general-distance-proof-absent} must be equal to
  \myeq{general-distance-proof-present}, and so 
  \begin{equation}
    \label{eq:general-distance-polynomial}
    0 = (p_{n+1})^n + n (p_{n+1})^{n-1} q_{n+1} (p_{n+1} + q_{n+1}) - (p_n)^n.
  \end{equation}

  Consider $n=2$.  By substituting the definitions of $p_2$, $p_3$, and $q_3$, and then rearranging terms, we may rewrite \myeq{general-distance-polynomial} as
  \begin{equation*}
    0 = \frac{\alpha f(0)^2  (2f(0)^2 - \alpha^2)}{(\alpha+f(0))^2 (\alpha+2f(0))^3},
  \end{equation*}
  which is satisfied only when $f(0) \in \{0, \alpha/\sqrt{2}\}$.  Consider the second of these roots, $\alpha/\sqrt{2}$.  When $n=3$, this value of $f(0)$ violates \myeq{general-distance-polynomial}.  Thus, the first root is the only possibility and we must have $f(0) = 0$.

\end{proof}

The decay function $f=0$ described in
Proposition~\ref{p:general-marginal-invariance} is a special case of
the decay function from
Proposition~\ref{p:sequential-marginal-invariance}, obtained by taking
$A=\emptyset$.  As described above, the resulting probability
distribution is one in which each customer links to himself, and is
thus clustered by himself.  This distribution is marginally invariant.
From this observation quickly follows the following corollary.

\begin{corollary}
  \label{c:general-not-marg}
  The decay function $f=0$ is the only one for which the resulting
  distance dependent CRP is marginally invariant over all distances,
  both sequential and non-sequential.  
\end{corollary}
\begin{proof}
Necessity of $f=0$ for marginal invariance follows from Proposition~\ref{p:general-marginal-invariance}.  Sufficiency follows from the fact that the probability distribution on partitions induced by $f=0$ is the one under which each customer is clustered alone almost surely, which is marginally invariant.
\end{proof}



\section{Gibbs sampling for the hyperparameters}
\label{app:hyperparams}

To enhance our models, we place a prior on the concentration parameter
$\alpha$ and augment our Gibbs sampler accordingly, just as is done in
the traditional CRP mixture~\citep{Escobar:1995}.  To sample from the
posterior of $\alpha$ given the customer assignments $\bc$ and
data, we begin by noting that $\alpha$ is conditionally independent of
the observed data given the customer assignments.  Thus, the quantity
needed for sampling is
\begin{equation*}
  p(\alpha \g \bc) \propto p(\bc\g \alpha) p(\alpha),
\end{equation*}
where $p(\alpha)$ is a prior on the concentration parameter.

From the independence of the $c_i$ under the generative process, 
$p(\bc\g \alpha) = \prod_{i=1}^N p(c_{i} \g D, \alpha)$.  Normalizing
provides
\begin{align*}
  p(\bc\g \alpha) &= \prod_{i=1}^N \frac{\ind{c_i=i}\alpha +
    \ind{c_i\ne
      i}f(d_{ic_i})}{\alpha + \sum_{j \neq i} f(d_{ij})}\\
  &\propto \alpha^{K} \left[\prod_{i=1}^N \left(\alpha +
      \sum_{j \neq i} f(d_{ij})\right)\right]^{-1},
\end{align*}
where $K$ is the number of self-links $c_i=i$ in the customer assignments $\bc$.
Although $K$ is equal to the number of tables $|z(\bc)|$ when distances are sequential, $K$ and $|z(\bc)|$
generally difffer when distances are non-sequential.
Then,
\begin{equation}
  \label{eq:posterior_alpha}
  p(\alpha \g \bc)
  \propto \alpha^{K} \left[\prod_{i=1}^N \left(\alpha +
      \sum_{j \neq i} f(d_{ij})\right)\right]^{-1} p(\alpha).
\end{equation}

\myeq{posterior_alpha} reduces further in the following special case: 
$f$ is the window decay function, $f(d) = \ind{d<a}$; 
$d_{ij}=i-j$ for $i>j$;
and distances are sequential so $d_{ij}=\infty$ for $i<j$.
In this case, $\sum_{j=1}^{i-1} f(d_{ij}) = (i-1)\wedge(a-1)$, where $\wedge$
is the minimum operator, and
 \begin{equation}
   \label{eq:window_term}
   \prod_{i=1}^N \left(\alpha + \sum_{j=1}^{i-1} f(d_{ij})\right) =
   (\alpha+a-1)^{[N-a]^+} \Gamma(\alpha+a\wedge N)/\Gamma(\alpha),
 \end{equation}
 where $[N-a]^+=\max(0,N-a)$ is the positive part of $N-a$.  Then,
 \begin{equation}
   \label{eq:posterior_alpha_window}
   p(\alpha \g \bc)
   \propto \frac{\Gamma(\alpha)}{\Gamma(\alpha+a\wedge
     N)}\frac{\alpha^{K}}{(\alpha+a-1)^{[N-a]^+}}p(\alpha).
 \end{equation}

 If we use the identity decay function, which results in the traditional CRP,
 then we recover an expression from \cite{Antoniak:1974}: $p(\alpha \g
 \bc) \propto
 \frac{\Gamma(\alpha)}{\Gamma(\alpha+N)}\alpha^{K}p(\alpha)$.  This
 expression is used in \cite{Escobar:1995} to sample exactly from the
 posterior of $\alpha$ when the prior is gamma distributed.

In general, if the prior on $\alpha$ is continuous then it is
difficult to sample exactly from the posterior of
\myeq{posterior_alpha}.  There are a number of ways to address this.
We may, for example, use the Griddy-Gibbs method~\citep{Ritter:1992}.
This method entails evaluating \myeq{posterior_alpha} on a finite set
of points, approximating the inverse cdf of $p(\alpha \g \bc)$
using these points, and transforming a uniform random variable with
this approximation to the inverse cdf.

We may also sample over any hyperparameters in the decay function
used (e.g., the window size in the window decay function, or the rate
parameter in the exponential decay function) within our Gibbs sampler.  For the rest of this
section, we use $a$ to generically denote a hyperparameter in the
decay function, and we make this dependence explicit by writing
$f(d,a)$.

To describe Gibbs sampling over these hyperparameters in the decay function,
we first write
 \begin{align*}
   p(\bc\mid \alpha,a)
   &= \prod_{i=1}^N \frac{\ind{c_i=i}\alpha + \ind{c_i\ne i}f(d_{ic_i},a)}{\alpha + \sum_{j=1}^{i-1} f(d_{ij},a)}\\
   &= \alpha^{K} \left[\prod_{i: c_i \ne i} f(d_{ij},a)\right]
   \left[\prod_{i=1}^N \left(\alpha + \sum_{j=1}^{i-1}
       f(d_{ij},a)\right)\right]^{-1}.
 \end{align*}

 Since $a$ is conditionally independent of the observed data given
 $\bc$ and $\alpha$, to sample over $a$ in our Gibbs sampler it is
 enough to know the density
 \begin{equation}
 \label{eq:posterior_a}
 p(a\mid \bc,\alpha)
 \propto
 \left[\prod_{i : c_i \ne i} f(d_{ij},a)\right]
 \left[\prod_{i=1}^N \left(\alpha + \sum_{j=1}^{i-1} f(d_{ij},a)\right)\right]^{-1}
 p(a\mid \alpha).
 \end{equation}
 In many cases our prior $p(a\mid\alpha)$ on $a$ will not depend on $\alpha$.

 In the case of the window decay function with sequential distances and $d_{ij}=i-j$ for $i>j$, we can
 simplify this further as we did above with \myeq{window_term}.  
 Noting that $\prod_{i: c_i \ne i} f(d_{ij},a)$ will be
 $1$ for those $a>\max_i i-c_i$, and $0$ for other $a$, we have
 \begin{equation}
   p(a\mid \bc,\alpha)
   \propto
   \frac{\Gamma(\alpha)}{\Gamma(\alpha+a\wedge N)}\frac{p(a\mid
     \alpha)\ind{a>\max_i i-c_i}}{(\alpha+a-1)^{[N-a]^+}}.
 \end{equation}

 If the prior distribution on $a$ is discrete and concentrated on a
 finite set, as it might be with the window decay function, one can
 simply evaluate and normalize \myeq{posterior_a} on this set.  If
 the prior is continuous, as it might be with the exponential
 decay function, then it is difficult to sample exactly from
 \myeq{posterior_a}, but one can again use the Griddy-Gibbs
 approach of \cite{Ritter:1992} to sample approximately.